\documentclass[journal]{IEEEtran}

\ifCLASSINFOpdf
\else
\fi
\usepackage{graphicx}
\usepackage{amsmath}
\usepackage{cite}

 \newcommand{\etal}{\textit{et al}.}
\newcommand{\ie}{\textit{i}.\textit{e}.}
\newcommand{\eg}{\textit{e}.\textit{g}.}

\begin{document}

\title{DR-Net: Transmission Steered Single Image Dehazing Network with Weakly Supervised Refinement}

\author{Chongyi~Li,
        Jichang~Guo,
        Fatih~Porikli,
        Chunle~Guo,
        Huzhu~Fu,
        Xi~Li 
\thanks{This work was supported in part by the National Key Basic Research Program of China (2014CB340403), the National Natural Science Foundation of China (61771334), the Natural Science Foundation of Tianjin of China (15JCYBJC15500), and the program of China Scholarships Council (CSC) under the Grant CSC No. 201606250063.}
\thanks{Chongyi Li is with the School of Electrical and Information Engineering, Tianjin University, Tianjin, China and Research School of Engineering, Australian National University, Canberra, ACT 0200, Australia (e-mail: lichongyi@tju.edu.cn).

Jichang Guo and Chunle Guo are with the School of Electrical and Information Engineering, Tianjin University, Tianjin, China (e-mail: jcguo@tju.edu.cn; guochunle@tju.edu.cn).

Fatih Porikli is with Research School of Engineering, Australian National University, Canberra, ACT 0200, Australia (e-mail: fatih.porikli@anu.edu.au).

Huazhu Fu is with Institute for Infocomm Research, at Agency for Science, Technology and Research, Singapore (e-mail: huazhufu@gmail.com).

Xi Li is with College of Computer Science and Technology, Zhejiang University, Hangzhou, Zhejiang, China (e-mail: xilizju@zju.edu.cn).
}}
\markboth{}%
{Shell \MakeLowercase{\textit{et al.}}: Bare Demo of IEEEtran.cls for Journals}

\maketitle

\begin{abstract}

Despite the recent progress in image dehazing, several problems remain largely unsolved such as robustness for varying scenes, the visual quality of reconstructed images, and effectiveness and flexibility for applications. To tackle these problems, we propose a new deep network architecture for single image dehazing called DR-Net. Our model consists of three main subnetworks: a transmission prediction network that predicts transmission map for the input image, a haze removal network that reconstructs latent image steered by the transmission map, and a refinement network that enhances the details and color properties of the dehazed result via weakly supervised learning.  Compared to previous methods, our method advances in three aspects: (i) pure data-driven model; (ii) the end-to-end system; (iii) superior robustness, accuracy, and applicability. Extensive experiments demonstrate that our DR-Net outperforms the state-of-the-art methods on both synthetic and real images in qualitative and quantitative metrics. Additionally, the utility of DR-Net has been illustrated by its potential usage in several important computer vision tasks.

\end{abstract}

\section{Introduction}

High-quality images are desired in computer vision applications and multimedia content sharing. However, images captured in outdoors environments often suffer from noticeable interference from the haze, which is a natural atmospheric phenomenon caused by floating particles (\eg, dust, smoke, and liquid droplets). Haze has two main effects on the captured images: attenuation of the light and contamination with an additive component to the image \cite{Berman2016}. Specifically, the scattering of floating particles distorts the direct transmission of light from the scene to the camera,   deviating the light from a straight trajectory and dispersing it if the particles are comparable in size to the wavelength of the probing light. The attenuated transmission decreases of intensity while the surrounding scattered light induces a blurred appearance of the scene \cite{Li2016survey}.


To remove the haze artifacts, previous single image dehazing methods usually follow the similar pipeline of (1) modeling the medium transmission, (2) refining the coarse transmission model, (3) estimating the global atmospheric light, and (4) reconstructing the latent image according to the predicted model parameters. However, this pipeline imposes several limitations. Firstly, transmission is ordinarily estimated based on priors. However, the priors relying on statistics are not accurate when hazy images captured under uncontrolled light conditions, different haze concentrations, and varying scene depth. Secondly, conventional global atmospheric light estimation methods (\eg, dark channel, quad-tree sub-division, bright channel, \etal) often make mistakes when there are white objects, highlight regions, or shadows. Thirdly, the errors in the separate estimation steps will be accumulated and amplified when the separately estimated variables are combined, which leads to suboptimal dehazing performance \cite{Li2017allinone}.

\begin{figure}[!t]
  \centering
\begin{minipage}[b]{0.3\linewidth}
  \centering
  \centerline{\includegraphics[width=2.5cm]{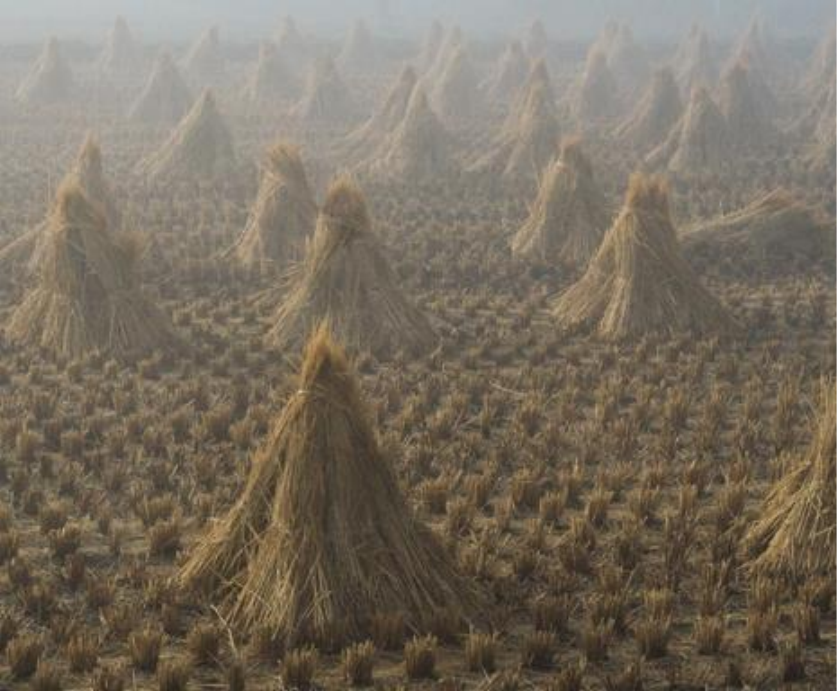}}
  \centerline{(a) Hazy image}\medskip
\end{minipage}
\begin{minipage}[b]{0.3\linewidth}
  \centering
  \centerline{\includegraphics[width=2.5cm]{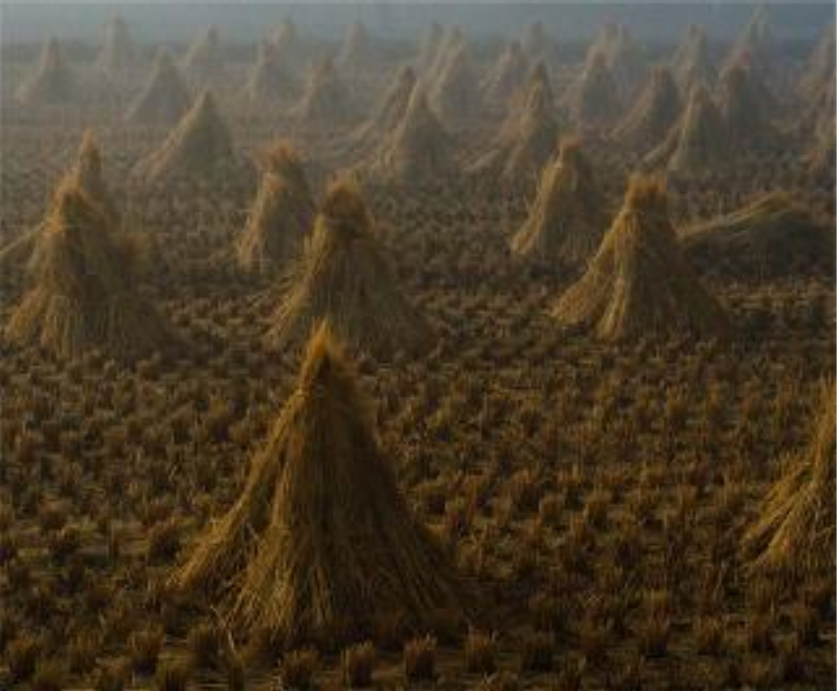}}
  \centerline{(b) \cite{Li2017allinone}}\medskip
\end{minipage}
\begin{minipage}[b]{0.3\linewidth}
  \centering
  \centerline{\includegraphics[width=2.5cm]{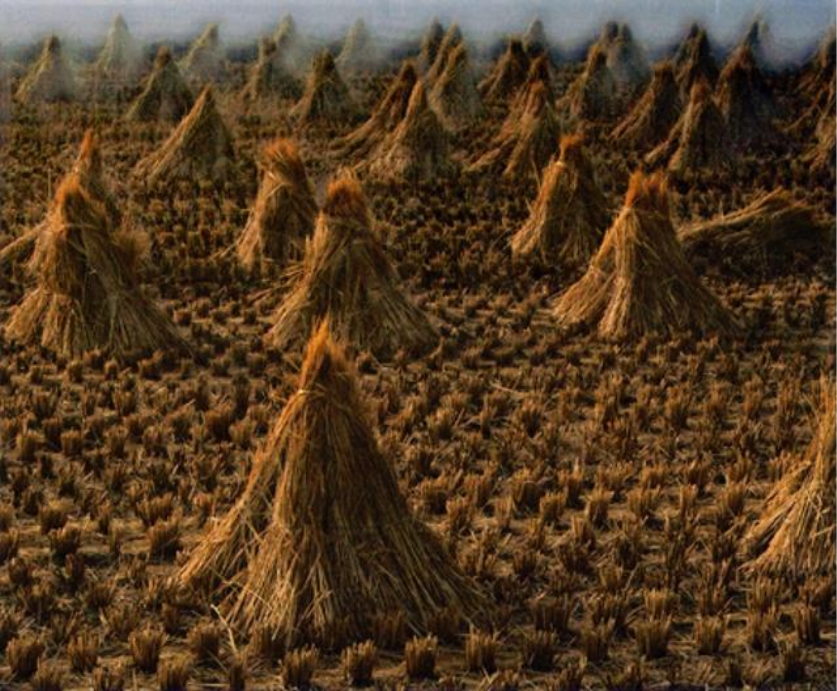}}
  \centerline{(c) Our result}\medskip
\end{minipage}
\caption{Sample dehazing result of DR-Net. }
\label{fig:example}
\vspace{-5mm}
\end{figure}

To tackle these limitations, we propose a new dehazing network that benefits from pure data-driven learning, end-to-end architecture, and weakly supervised refinement. Observing in Figure~\ref{fig:example}(a), it is obvious that the presence of haze greatly impairs the visual quality of the image. Compared to the result of the recent deep learning based method \cite{Li2017allinone} shown in Figure~\ref{fig:example}(b), our latent image (\ie, final refined result) has better contrast, details, and color.


\section{Related Work}

Image dehazing aims to recover a clear image from an image captured under hazy scenes. Many approaches developed to address this ill-posed problem can be categorized as supplementary-information based methods and single-image based methods.


Supplementary information based methods usually require additional knowledge such as 3D geographical models \cite{Kopf2008}, scene depth \cite{Tan2000}, multiple images of the scene under different weather conditions \cite{Narasimhan2003}, polarization filters \cite{Schechner2001}, and so forth. Nevertheless, these methods are mostly computationally intensive and not applicable to dynamic scenes. Much attention, therefore, has been devoted to single image dehazing methods \cite{Tan2008,Fattal2008,He2011,Meng2013,Ancuti2013, Fattal2014,Tang2014, Zhu2015, Berman2016, Cai2016, Ren2016}. In \cite{He2011}, He \etal observed an interesting phenomenon of haze-free outdoors images that at least one channel has some pixels with very low intensities. Based upon this prior, the transmission and global atmospheric light were roughly estimated. After that, the dehazed image was achieved based on the refined transmission by soft-matting \cite{Levin2006} or guided filter \cite{He2013} as well as the estimated global atmospheric light. In \cite{Berman2016}, Berman \etal proposed a nonlocal image dehazing method, which relied on the assumption that colors of a haze-free image are well approximated by a few hundred distinct colors that form tight clusters in RGB space.


With the emergence of deep learning solutions \cite{ LeCun2015}, deep learning based methods have achieved a promising performance in image dehazing. In \cite{Cai2016, Ren2016},  a convolutional neural network (CNN) was utilized to predict the transmission. After that, guided filtering \cite{He2013} as post-processing was used to suppress halo effect in the predicted transmission caused by the patch based prediction. With the transmission and the global atmospheric light estimated by conventional methods, the haze-free image was reconstructed. Different from Cai \etal \cite{Cai2016} and Ren \etal \cite{Ren2016}, which separately estimated model parameters and used post-processing, DR-Net directly produces a clear image in its end-to-end system. In \cite{Li2017allinone}, Li \etal estimated the parameters of a haze image formation model in one unified CNN model. Such all-in-one model made it easy to embed the model into other deep models.  Though deep learning based methods have advantages, there is still much room for improvement, such as robustness for varying scenes, fidelity and visual quality of reconstructed images, and flexibility for applications.


\subsection{Our Contributions}

DR-Net is a pure data-driven, end-to-end, fully-convolutional network that consists of three main subnetworks designed for specific tasks. The contributions are summarized as follows:
\begin{itemize}

\item To the best of our knowledge, this is the first attempt that investigates the combination of strongly and weakly supervised learning for single image dehazing. DR-Net reconstructs latent images based on strongly supervised learning, while the dehazed results are further refined by weakly supervised learning built on a Generative Adversarial Network (GAN). Furthermore, DR-Net includes a transmission prediction subnetwork that improves haze removal performance and training convergence.

\item Instead of following the traditional pipeline, DR-Net directly predicts clear images in a pure data-driven and end-to-end manner, which is more flexible and suitable for practical applications. In addition, DR-Net directly minimizes the reconstruction loss to avoid the accumulated errors from individual estimations of transmission and global atmospheric light. This produces more accurate reconstruction results.

\item DR-Net achieves the best performance on both synthetic and real hazy images. Additionally, DR-Net generalizes well to varying scenes and lighting conditions.
Code and data will be available after publication.
\end{itemize}

\section{DR-Net}

\begin{figure*}[htb]
 \centering
 \centerline{\includegraphics[width=15cm,height=9.2cm]{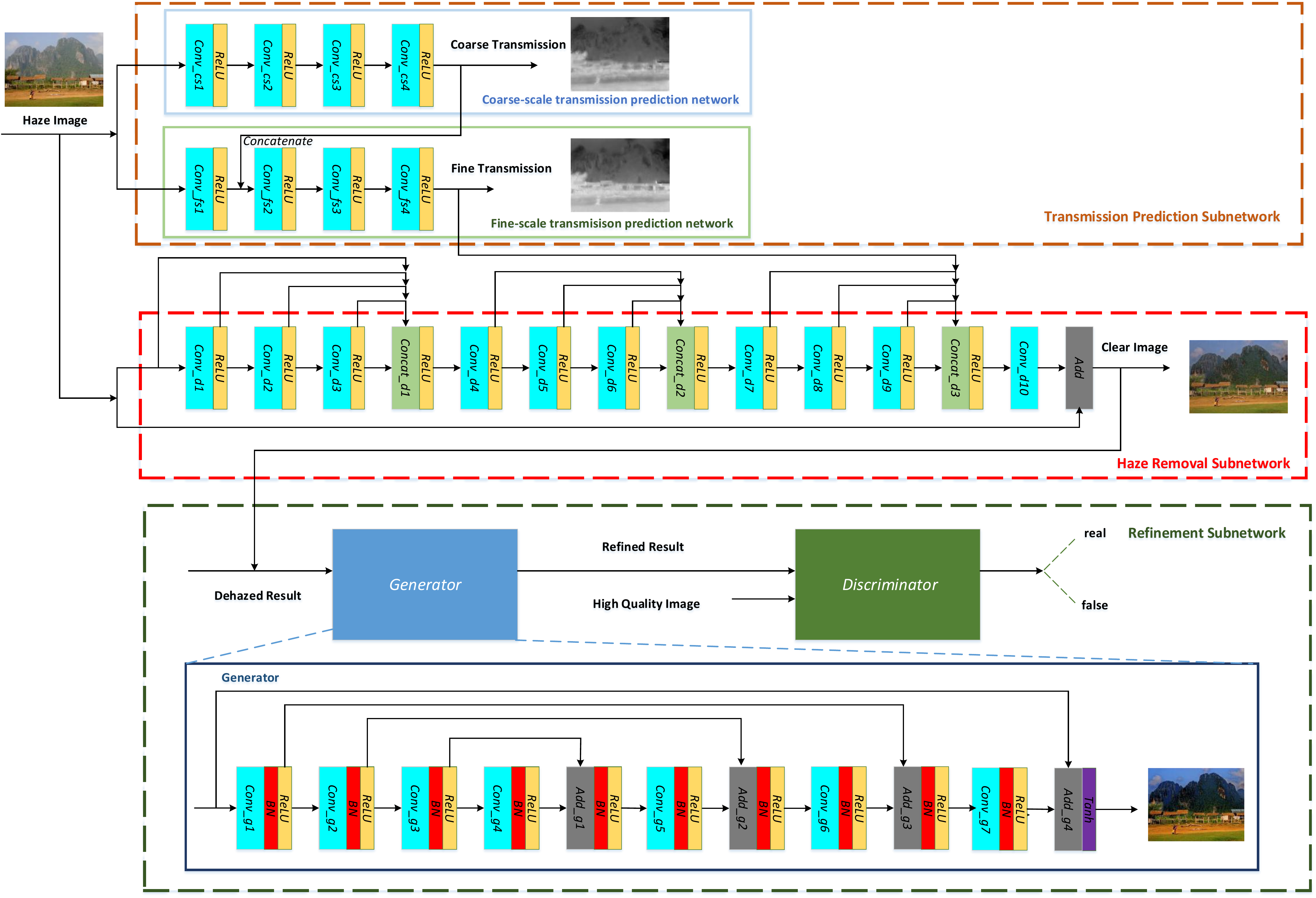}}
 \caption{ An overview of the DR-Net architecture. From top to the bottom are the transmission prediction subnetwork, haze removal subnetwork,
 and refinement subnetwork. Different color blocks represent different operations.}
 \label{fig:networks}
 \end{figure*}

To automatically reveal the underlying correlations between hazy image and haze-free image in an end-to-end manner, DR-Net employs a transmission prediction subnetwork, a haze removal subnetwork, and a refinement subnetwork. An overview of the DR-Net architecture is shown in Figure~\ref{fig:networks}. In what follows, we explain these three subnetworks in detail. Before that, we first formulate the problem.


\subsection{Problem Formulation }
For an image captured under the haze, only a part of the reflected light from the scene reaches the imaging sensor due to the absorption and scattering effects, which decrease the visibility and contrast of the scene. According to the atmospheric scattering model \cite{Koschmieder1924}, hazy image formation can be described as
 \begin{equation}
\label{equ_1}
I(x)=J(x)t(x)+A(x)(1-t(x)),
\end{equation}
where $x$ denotes the pixel coordinates, $I$ is the observed image, $J$ is the haze-free latent image, $A$ is the global atmospheric light, and $t(x) \in [0,1]$ is the transmission, which represents the percentage of the scene radiance reaching the camera. When the haze is homogenous, $t(x)$ can be further expressed in an exponential decay term as
\begin{equation}
\label{equ_2}
t(x)=\exp(-\beta d(x)).
\end{equation}
where $\beta$ is the atmospheric attenuation coefficient and $d(x)$ is the distance from the scene to the camera. The purpose of single image dehazing is to reconstruct $J$ from $I$.

%



\subsection{ Transmission Prediction Subnetwork }

Inspired by \cite{Zhang2017JointTM}, we employ a transmission prediction subnetwork in our DR-Net. This subnetwork also adopts the multi-scale fully convolutional network architecture proposed in \cite{Ren2016}, which consists of a coarse-scale network and a fine-scale network. The task of the coarse-scale network is to predict a holistic transmission of the scene and the task of the fine-scale network is to refine the textures of the transmission. In the fine-scale network, the outputs of the first layer are concatenated with the output from the coarse-scale network as the inputs of the second layer.
Unlike \cite{Ren2016}, we remove the pooling and up-sampling operations that tend to cause blurring in the predicted transmission output.

\subsubsection{Loss functions of the transmission prediction subnetwork}


For the coarse-scale transmission prediction network, we impose a reconstruction objective, that is, we minimize the Mean Squared Error (MSE) loss function
\begin{equation}
\label{equ_3}
L_{{cs}_{MSE}}=\frac{1}{NHW}\sum_{i=1}^{N}\|F_{cs}(X_{i})-t_i\|^{2},
\end{equation}
where $N$ is the number of each batch, $H\times W$ is the dimension of the transmission map, $F_{cs}$ is the learned transmission prediction mapping function of the coarse-scale network, $X_{i}$ is input hazy image, $F_{cs}(X_{i})$ is the predicted coarse transmission map, $t_i$ is the ground truth of transmission map.
Using MSE loss, a predicted transmission map with coarse details and textures is achieved. Then, the coarse transmission map is concatenated with the outputs of the first layer of fine-scale network as the second layer inputs of the fine-scale network.
For the fine-scale transmission prediction network, we also minimize the MSE loss function
\begin{equation}
\label{equ_5}
L_{{fs}_{MSE}}=\frac{1}{NHW}\sum_{i=1}^{N}\|F_{fs}(X_{i})-t_i\|^{2},
\end{equation}
where $F_{fs}$ is the learned transmission prediction mapping function of the fine-scale network.

To further preserve structure and texture of the predicted fine transmission, we add a structural similarity index (SSIM) loss \cite{Zhao2017} to fine-scale network. Firstly, the SSIM value for every pixel between the predicted fine transmission map $F_{fs}(X_{i})$ and the ground truth $t_i$ is calculated as follows:
\begin{equation}
\label{equ_6}
SSIM(p)=\frac{2\mu_{x}\mu_{y}+C_{1}}{\mu_{x}^{2}+\mu_{y}^{2}+C_{1}}\cdot\frac{2\sigma_{xy}+C_{2}}{\sigma_{x}^{2}+\sigma_{y}^{2}+C_{2}},
\end{equation}
where $x$ and $y$ are the corresponding image patches with size $13\times13$ in the fine transmission map and ground truth, respectively. Above, $p$ is the center pixel of image patch, $\mu_{x}$ is the mean of $x$, $\sigma_{x}$ is the standard deviations of $x$, $\mu_{y}$ is the mean of $y$, $\sigma_{y}$ is the standard deviations of $y$,  $\sigma_{xy}$ is the covariance between $x$ and $y$. Using the parameters default in the SSIM loss, we set the values of $C_{1}$ and $C_{2}$ to 0.02 and 0.03.
Using \eqref{equ_6}, the SSIM loss between the predicted fine transmission map and the ground truth transmission map is expressed as
\begin{equation}
\label{equ_7}
L_{{fs}_{SSIM}}=\frac{1}{N}\sum_{i=1}^{N}(1-\frac{1}{M}\Sigma_{p=1}^{M}(SSIM(p))),
\end{equation}
where $M=H\times W$ is the dimension of transmission map.

The final loss function for the transmission prediction subnetwork is the linear combination of the above-introduced losses with the following weights:
\begin{equation}
\label{equ_8}
\begin{aligned}
L_{{tp}_{total}}=L_{{cs}_{MSE}}+L_{{fs}_{MSE}}+L_{{fs}_{SSIM}}.
\end{aligned}
\end{equation}
The blending weights are picked empirically based on preliminary experiments on the training data.

\subsection{Haze Removal Subnetwork}

Transmission represents the percentage of the scene radiance reaching the camera and transmission map indicates the haze concentrations in the input hazy image, which is the significant clue for haze removal. Different from previous methods that estimate transmission map and then directly reconstruct latent result based on the guidance of transmission map, we feed the predicted fine transmission map to the haze removal subnetwork as an additional feature map in order to automatically reveal the underlying correlations among hazy image, transmission map, and haze-free image.

For the haze removal subnetwork, we learn the residual between the original hazy image and the corresponding haze-free image. Residual learning  allows the end-to-end training easier and more effective, because this may simply drive the weights of multiple nonlinear layers toward zero \cite{He2016}. Although the concept of predicting residuals has been used in previous methods \cite{Kim2016, Fu2017, Zhang2017cvpr, Zhang2017denoising}, it has not been studied in the context of learning based haze removal. Furthermore, we boost it by stacking early layers at the end of each block, which strengthens feature propagation and alleviates the vanishing-gradient problem \cite{Huang2017}. The haze removal subnetwork architecture can be found in Figure~\ref{fig:networks}.

\subsubsection{Loss functions of the haze removal subnetwork}
To learn the residual mapping, we add the estimated residual to the input hazy image and minimize the MSE loss
\begin{equation}
\label{equ_9}
L_{d_{MSE}}=\frac{1}{NCHW}\sum_{i=1}^{N}\|(F_d(X_{i})+X_i)-Y_i\|^{2},
\end{equation}
where $F_{d}$ is the learned residual mapping function, $Y_{i}$ is the ground truth of hazy image $X_{i}$, and $C\times H\times W$ is the dimension of the input image.  However, we note that a direct optimizing MSE loss function tends to introduce artifacts and fake boundaries. Thus, similar to our formulation in \eqref{equ_6} and \eqref{equ_7}, we also compute the SSIM loss ($L_{d_{SSIM}}$) between the dehazed result and the ground truth. The final loss function for the haze removal subnetwork is the linear combination of the aforementioned losses:
\begin{equation}
\label{equ_10}
\begin{aligned}
L_{{d}_{total}}=&L_{d_{MSE}}+L_{d_{SSIM}}. \end{aligned}
\end{equation}
Our haze removal network is a fully convolutional network for computational efficiency and it does not include batch normalization layers to not discard any useful image details.

%

\subsection{Refinement Subnetwork}

The combination of the transmission prediction subnetwork and haze removal subnetwork can yield a clear image, however this dehazed output might have imprecise color range and contrast, in particular for outdoors data. The reason is that the dehazing subnetwork is trained with synthetic hazy images using an indoor RGB-D dataset.

To remedy this problem, we design a refinement subnetwork to enhance the contrast and color of the dehazed output based on weakly supervised learning. The refinement network is inspired by weakly supervised learning models \cite{Kim2017, Deng2017, Ignatov2017} which aim at capturing special characteristics of a given image collection and modeling how these characteristics could be translated onto another image collection.

Our goal is to learn a mapping function from a source domain $X$ (\ie, low quality images with monotonous color and blur details) to a target domain $Y$ (\ie, high quality images with vivid color and clear details). Following the GAN concept, the refinement subnetwork consists of a generator G and a discriminator D. The task of G is to trick D so that D confuses the G's outputs (\ie, refined images) as high quality images. Specifically, G aims to refine the given low quality image while D attempts to distinguish whether the refined image is authentic or not. Note that, unlike \cite{Goodfellow2014,Radford2015,Arjovsky2017} that generate novel images, our network only enhances its input.

We use a fully convolutional network as the generator where we incorporate nested shortcut connections (\ie, skip connections) across the symmetric layers with the aim of addressing the vanishing gradient problem because shortcut connections can effectively propagate gradient in the process of back propagating. All convolutional layers in the generator network are followed by batch normalization and ReLU activation function, except for the last one, where a scaled hyperbolic tangent is applied to output.
We adopt a CNN based architecture proposed in \cite{Radford2015} as our discriminator due to its simplicity and effectiveness. The refinement subnetwork architecture can be found in Figure~\ref{fig:networks}.

\subsubsection{Loss functions of the refinement subnetwork}

For the generator function G: $X$ $\rightarrow$ $Y$ and the discriminator D, the adversarial loss is expressed as:
\begin{equation}
\label{equ_11}
\begin{aligned}
L_{GAN}(G,D,X,Y)&=E_{y\sim p_{data}(y)}[log D(y)]\\
&+E_{x\sim p_{data}(x)}[log(1-D(G(F_{c}(x))].
\end{aligned}
\end{equation}
where $x\in$ domain $X$, $y\in$ domain $Y$, $F_{c}$ is the learned haze removal mapping function, G tries to generate image $G(F_{c}(x))$ that looks similar to image from domain $Y$ while D aims to distinguish between $G(F_{c}(x))$ and real sample $y$. G tries to minimize this loss against an adversarial D that tries to maximize it. Besides, we expect that our refined result keeps the content and structure of the dehazed result. Thus, we include MSE and SSIM losses to the refinement subnetwork optimization. The MSE loss is expressed as
\begin{equation}
\label{equ_12}
L_{{rf}_{MSE}}=\frac{1}{NCHW}\sum_{i=1}^{N}\|G(F_{c}(x_{i}))-F_{c}(x_{i})\|^{2},
\end{equation}
where $N$ is the number of each batch, $F_{c}$ is the learned haze removal mapping function, $G$ is the learned mapping function of the generator network, $x_{i}$ is input hazy image, $C\times H\times W$ is the dimension of the input image. The calculation of SSIM loss $L_{{rf}_{SSIM}}$ is similar to \eqref{equ_6} and \eqref{equ_7}. The total loss for the refinement subnetwork is expressed as
\begin{equation}
\label{equ_13}
L_{{rf}_{total}}=L_{{rf}_{MSE}}+L_{{rf}_{SSIM}}+10^{-3}L_{GAN}.
\end{equation}
The weights are fine-tuned on the training data.

\subsection{Training and Implementation Details}

\subsubsection{DR-Net Training}

Instead of training the transmission prediction subnetwork, and then using the transmission output to train the haze removal subnetwork, we optimize these two subnetworks at the same time. To stabilize our training, we adopt stage-wise learning scheme for refinement subnetwork, which separately optimizes the dehazing subnetworks (\ie, the transmission prediction and haze removal subnetworks together) and refinement subnetwork.

To train the dehazing subnetworks, we first synthesize a hazy image dataset according to \eqref{equ_1} and \eqref{equ_2} using RGB-D images from NYU Depth dataset \cite{Silberman2012}. Here, we assume that each channel of an image has the same global atmospheric light value and transmission value. Then, we randomly select global atmospheric light $A$ from [0.7, 1.0] and set the atmospheric attenuation coefficient $\beta$ varying from 0.6 to 1.6. We divide NYU Depth dataset into two parts: training part with 1299 RGB-D images and validation part with 150 RGB-D images. For each RGB-D image, we randomly select 20 global atmospheric light and atmospheric attenuation coefficient values to synthesize 20 hazy images and 20 corresponding transmission maps. Last, we obtain a training set including $1299\times 20$ training samples and a validation set including $150\times 20$ validation samples. Those synthetic samples include hazy images with different haze concentrations and light intensities and the corresponding transmission maps. We resize these samples to size $310\times 230$.

For the refinement subnetwork training, we first select 3000 high quality images from CUHK-Photo Quality Dataset \cite{Tang2013}. Next, we download 3000 hazy images from Internet, and then process them using our dehazing subnetworks. The dehazed results are used as low quality images. With the 3000 image pairs with size $640\times 480$, we update the generator G and the discriminator G in a alternating manner. For the generator, we first train it using MSE and SSIM losses. After converging, we add adversarial loss to it for preserving the content and structure of original images.

\subsubsection{Implementation}

DR-Net was implemented on a computer with Nvidia GTX 1080Ti GPU, Intel I7 CPU 4.0GHz and 32GB RAM using the TensorFlow framework. We trained the DR-Net using ADAM \cite{Kingma2014} and set the learning rate and momentums to 0.0002 and 0.9 for these three subnetworks. The batch sizes for the dehazing subnetworks and the refinement subnetwork were 16 and 8, respectively. It took 10 hours to optimize entire DR-Net (the dehazing subnetworks about 6 hours; refinement subnetwork about 4 hours). Note that the optimization of DR-Net is very fast. The reasons are 1) transmission map as an additional feature map accelerates training convergence; 2) the architecture of DR-Net is simple and efficient.
The processing time of transmission prediction subnetwork, haze removal subnetwork, and refinement subnetwork for an image with size $640\times 480$ are 0.04s, 0.06s and 0.15s (\ie, total time is 0.25s) on the above-mentioned machine. In contrast to previous methods that rely on guided filtering to suppress artifacts, our method directly produces clear results without relying on post-processing, which significantly reduces processing time. DR-Net is very fast for practical use.

\subsubsection{Parameter settings}

Transmission prediction subnetwork: For the coarse-scale, from the first layer to the fourth layer consist of 16, 16, 16, and 1 filters with size $11\times 11$, $9\times 9$, $7\times 7$, and $5\times 5$, respectively. For the fine-scale, from the first layer to the fourth layer consist of 16, 16, 16, and 1 filters with size $7\times 7$, $5\times 5$, $3\times 3$, and $1\times 1$, respectively.

\noindent Haze removal subnetwork: We use $3\times3$ filter size. The number of feature map for each convolutional layer is the same (\ie, 32) except for the last layer (\ie, 3).

\noindent Refinement subnetwork: For the generator network, all of convolutional layers have 32 filters with size $3\times 3$,  except for the last one.

In our architecture, we pad zeros for all convolution layers to preserve image size.
Additionally, our DR-Net is a fully convolutional network, which can be seamlessly embedded with other deep network architectures.

We also investigated the impacts of kernel size, filter number and network depth on DR-Net. Similar to the conclusions of other low-level networks, the larger kernel sizes, more filters and deeper network depth can improve the performance to some extent at the cost of running time. Our DR-Net can generate good enough results with the above-mentioned parameters while the complexity and computation time can be reduced further as needed.

\section{Experiments}

To evaluate our DR-Net, we use both synthetic and real data and compare with several recent state-of-the-art single image dehazing methods: Meng \etal \cite{Meng2013}, Cai \etal \cite{Cai2016}, Ren \etal \cite{Ren2016}, and Li \etal \cite{Li2017allinone}. Among these methods, Cai \etal \cite{Cai2016}, Ren \etal \cite{Ren2016}, and Li \etal \cite{Li2017allinone} are deep learning based methods that are similar with our method. To further demonstrate the performance of our DR-Net, we also present the results of our dehazing subnetworks. The results after our dehazing subnetworks and refinement subnetwork are denoted as \textbf{ID} and \textbf{IR}, respectively.
More results are provided in the supplementary material.

To validate the generalization and utility of our method, we conduct experiments on challenging data and illustrate the potential usage of DR-Net in keypoint matching and object localization as well. We analyze the effects of transmission prediction subnetwork at last.

We do not compare the processing time since different methods are implemented on different platforms with or without GPU acceleration.

\subsection{Experiments on Synthetic Data}

\begin{figure*}[htb]
  \centering
\begin{minipage}[b]{0.11\linewidth}
  \centering
  \centerline{\includegraphics[width=1.9cm,height=3.5cm]{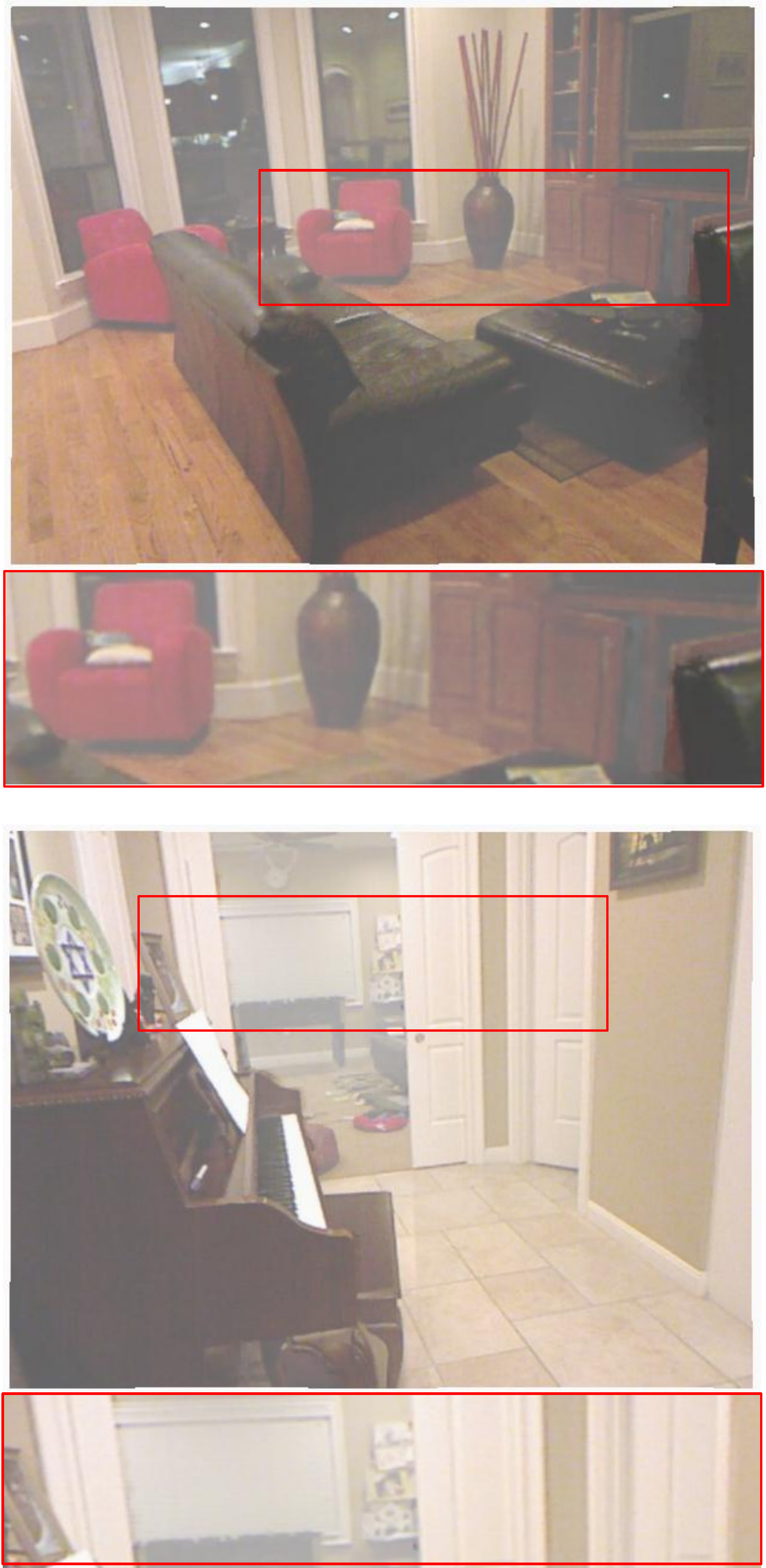}}
  \centerline{(a)Hazy images}\medskip
\end{minipage}
\begin{minipage}[b]{0.11\linewidth}
  \centering
  \centerline{\includegraphics[width=1.9cm,height=3.5cm]{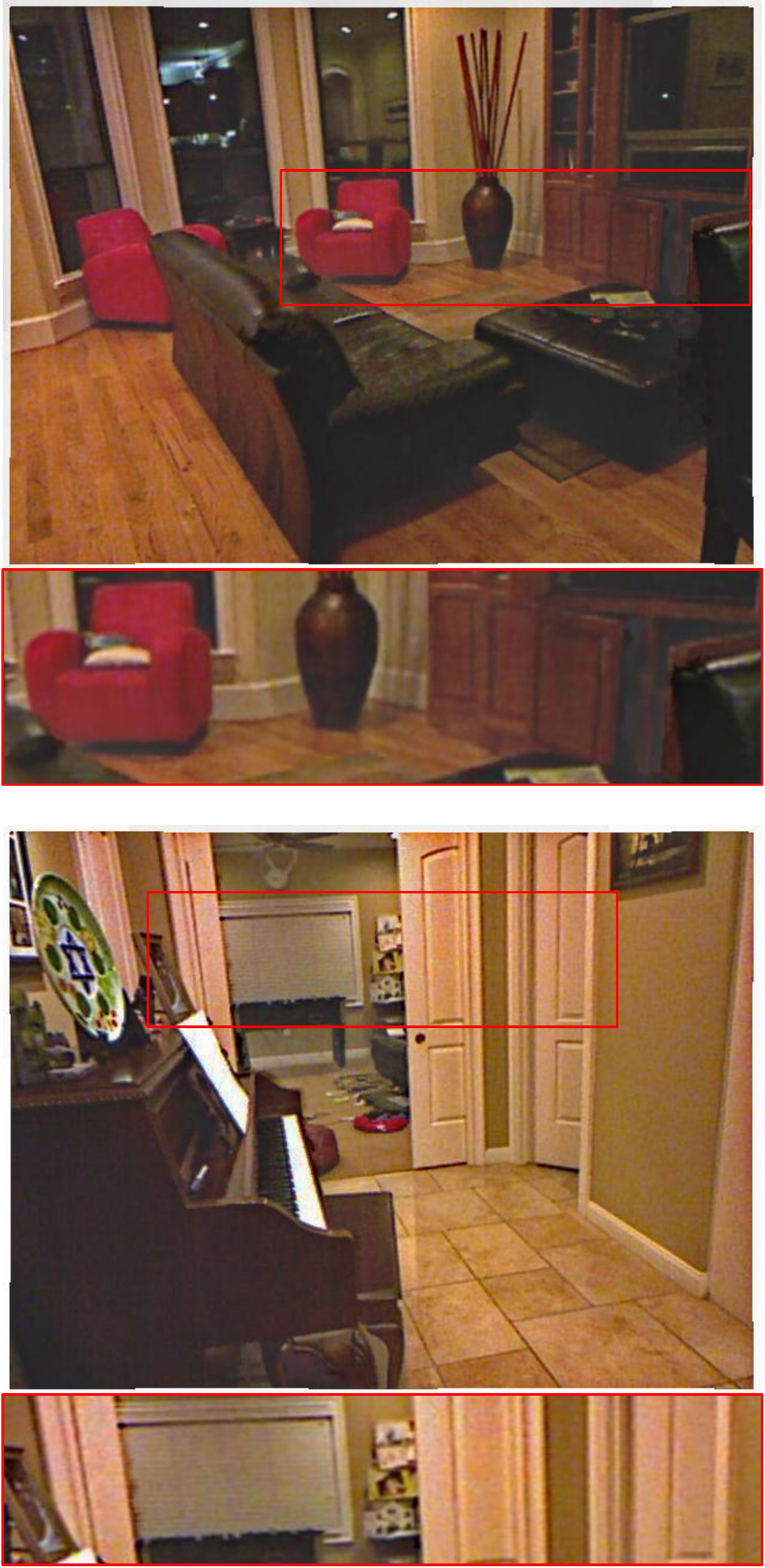}}
  \centerline{(b) \cite{Meng2013}}\medskip
\end{minipage}
\begin{minipage}[b]{0.11\linewidth}
  \centering
  \centerline{\includegraphics[width=1.9cm,height=3.5cm]{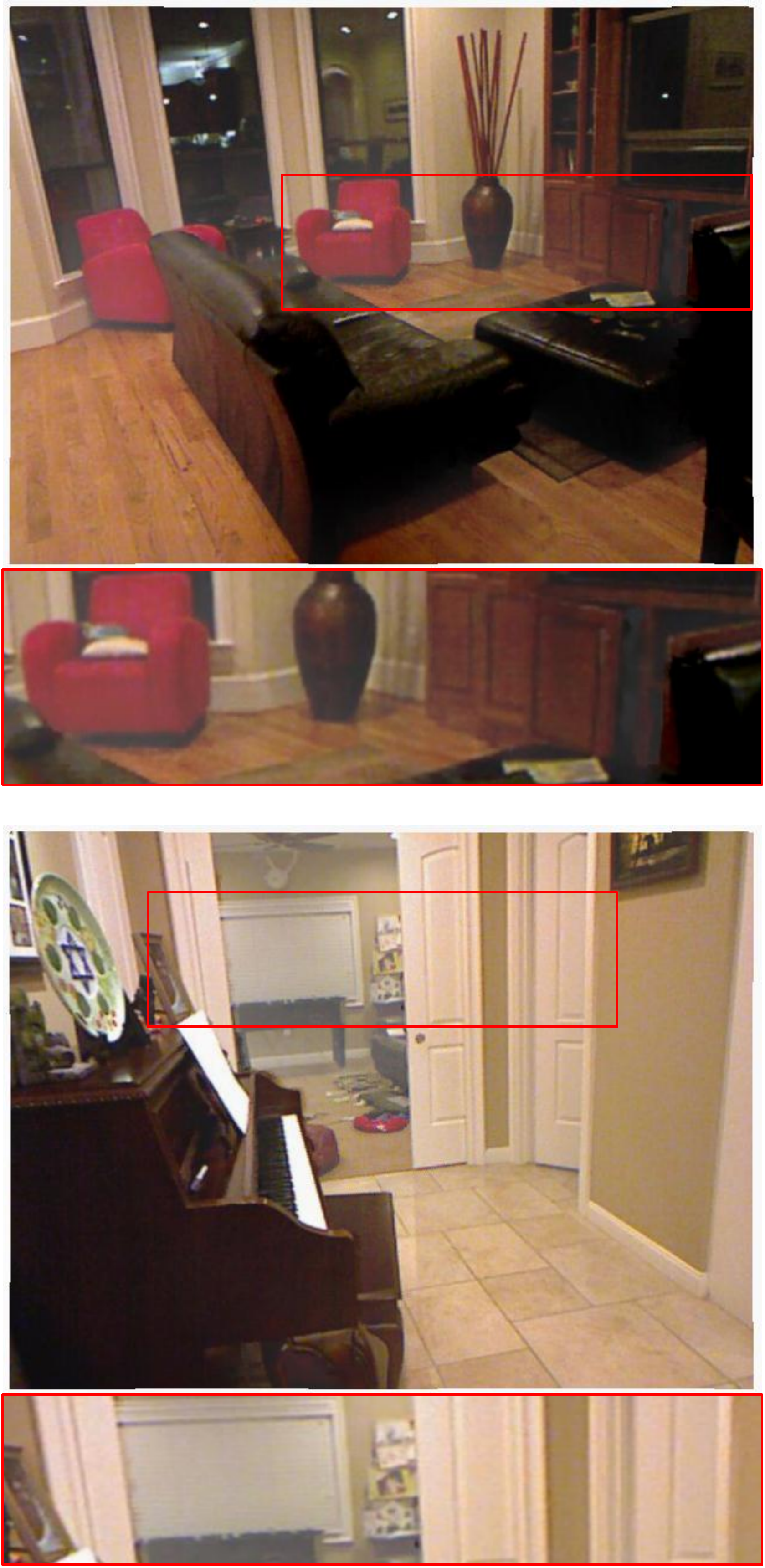}}
  \centerline{(c)   \cite{Cai2016}}\medskip
\end{minipage}
\begin{minipage}[b]{0.11\linewidth}
  \centering
  \centerline{\includegraphics[width=1.9cm,height=3.5cm]{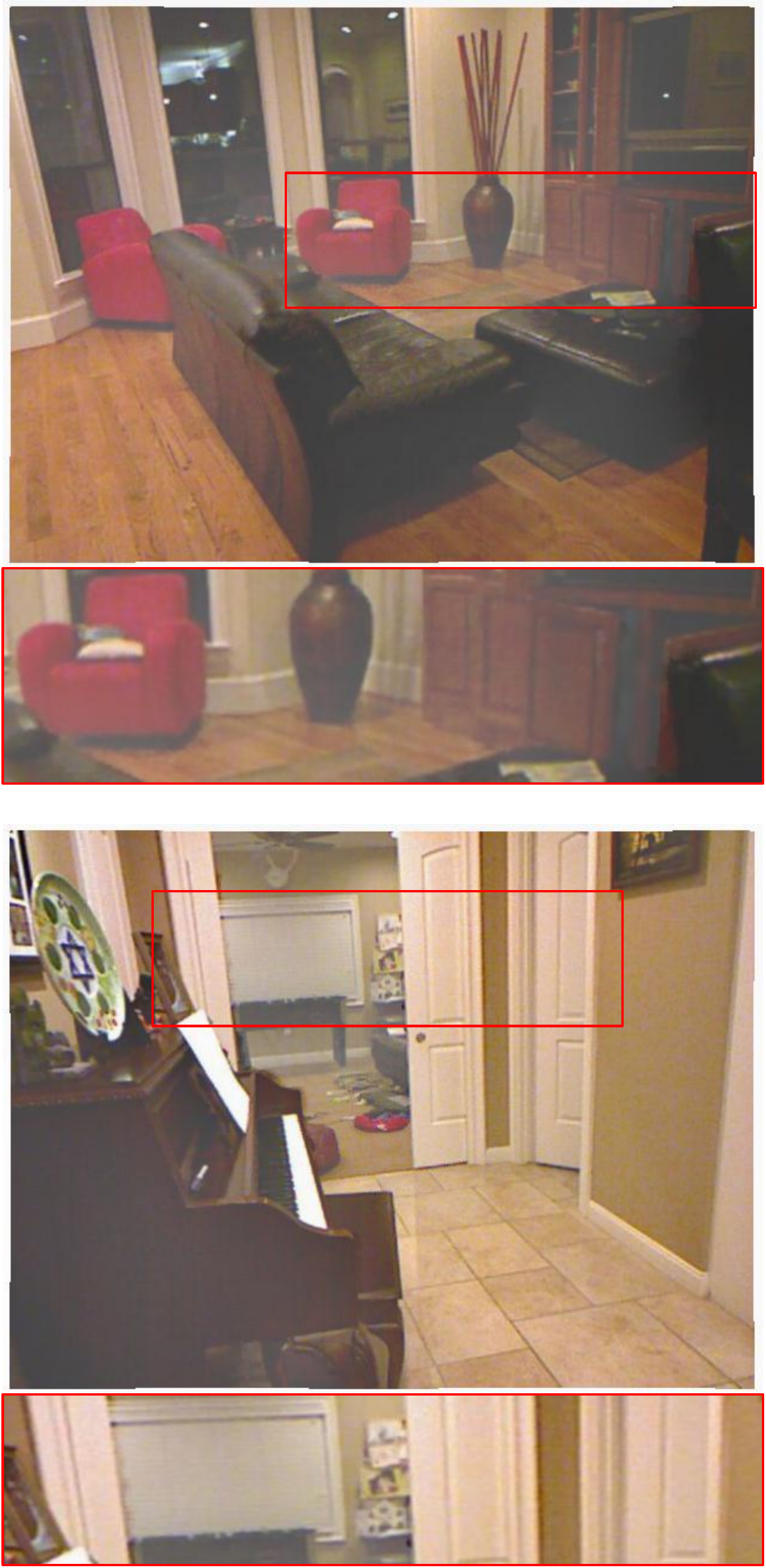}}
  \centerline{(d)   \cite{Ren2016}}\medskip
\end{minipage}
\begin{minipage}[b]{0.11\linewidth}
  \centering
  \centerline{\includegraphics[width=1.9cm,height=3.5cm]{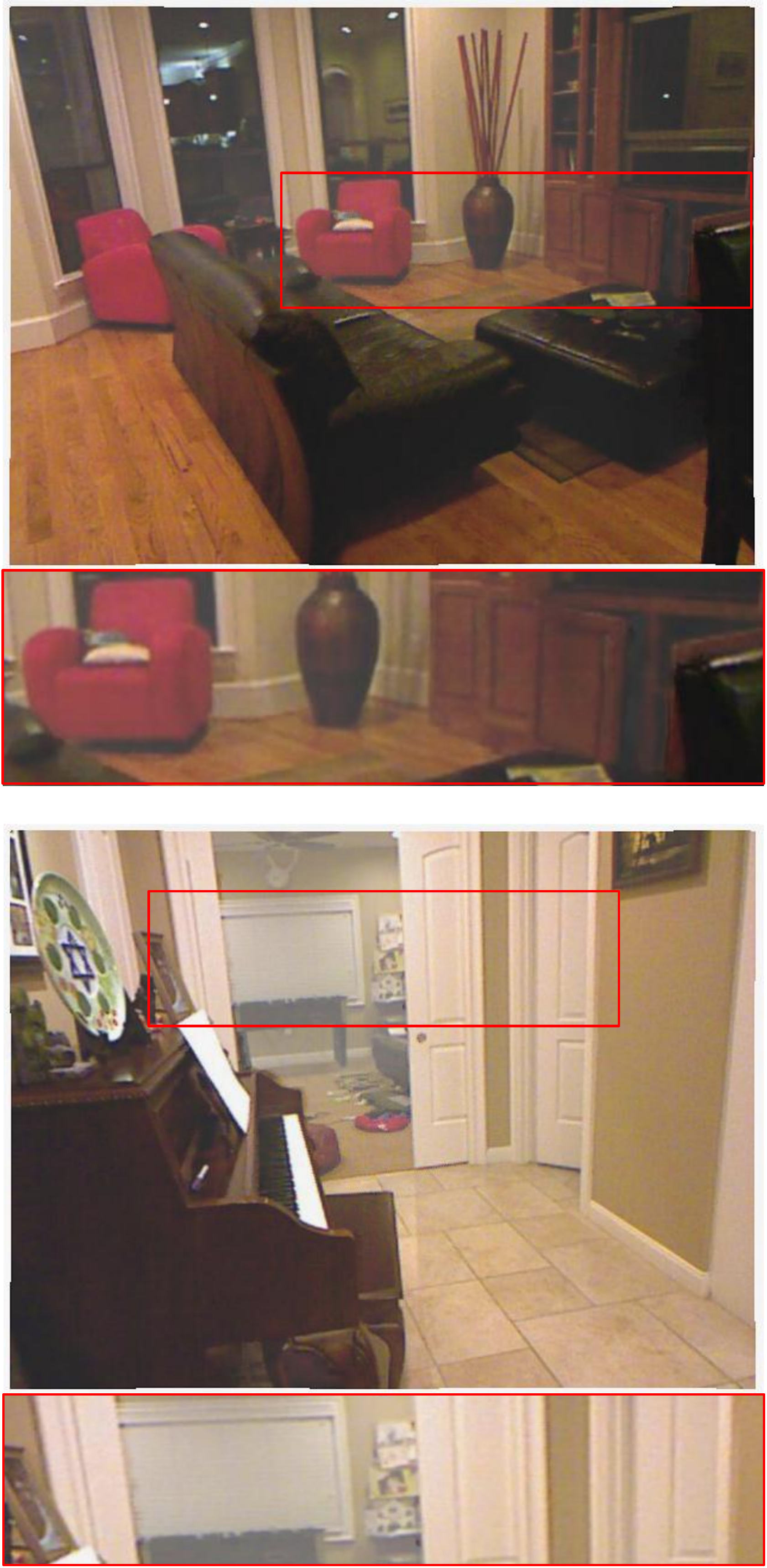}}
  \centerline{(e) \cite{Li2017allinone}}\medskip
\end{minipage}
\begin{minipage}[b]{0.11\linewidth}
  \centering
  \centerline{\includegraphics[width=1.9cm,height=3.5cm]{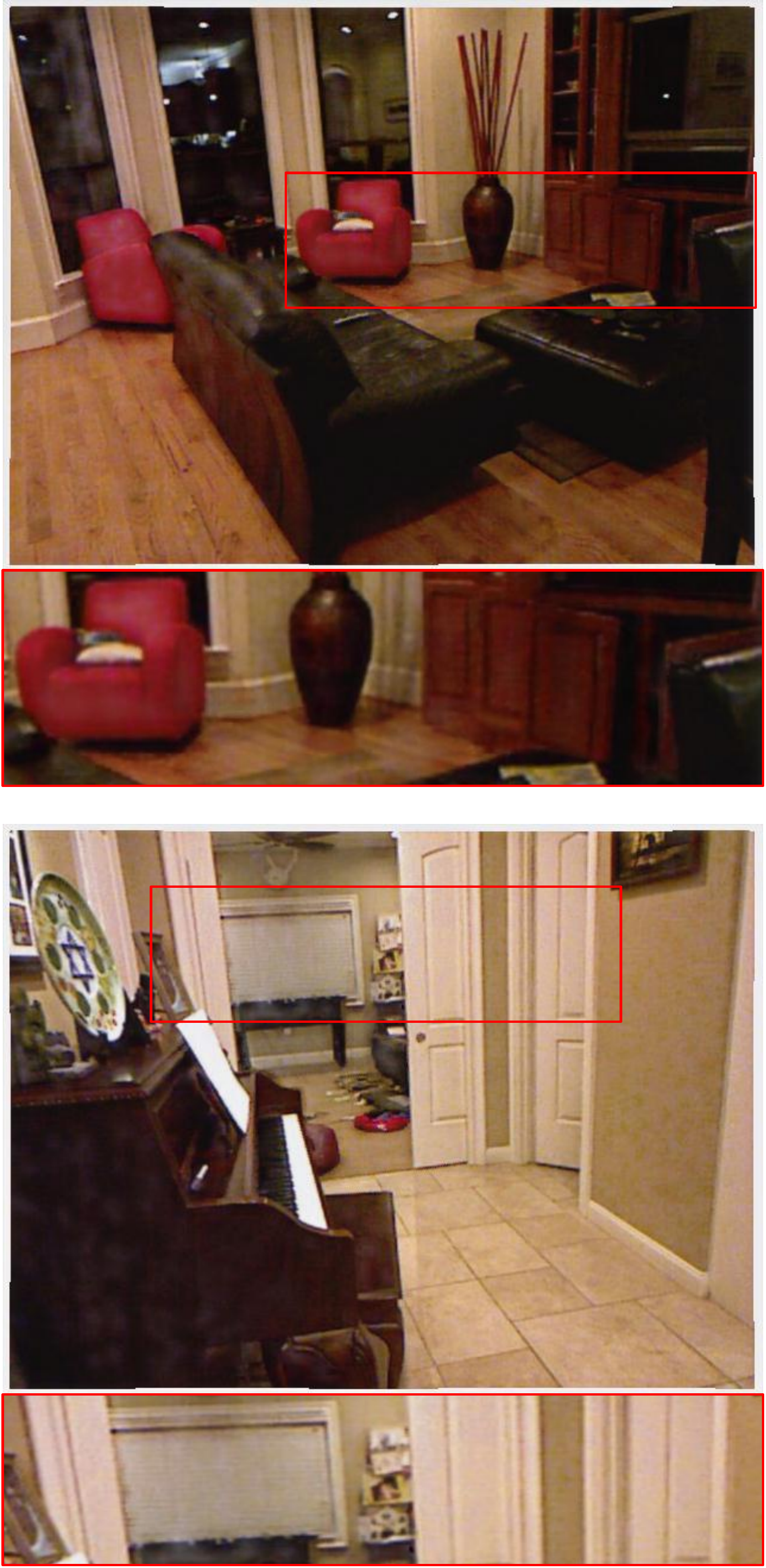}}
  \centerline{(f) ID }\medskip
\end{minipage}
\begin{minipage}[b]{0.11\linewidth}
  \centering
  \centerline{\includegraphics[width=1.9cm,height=3.5cm]{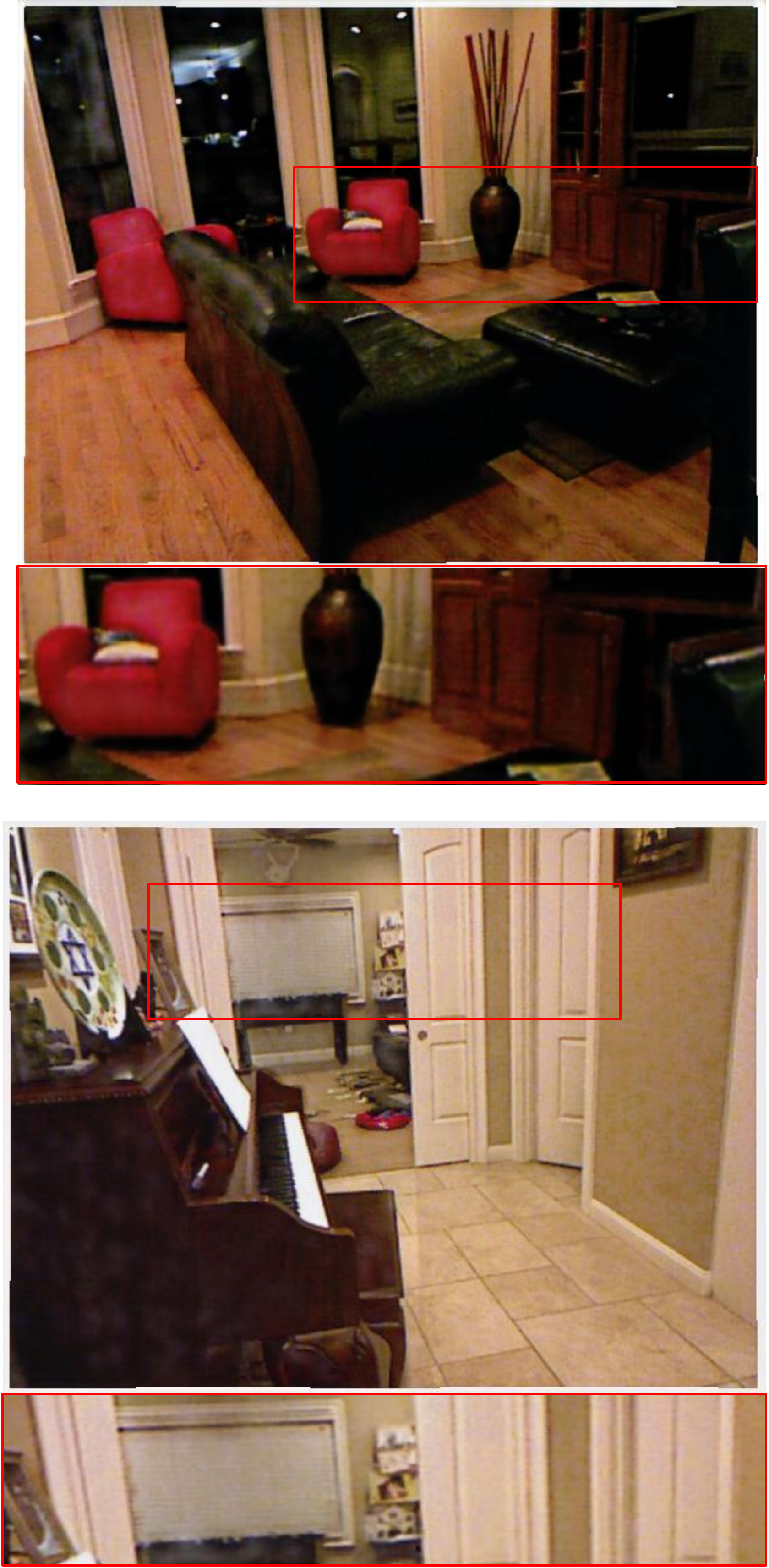}}
  \centerline{(g) IR }\medskip
\end{minipage}
\begin{minipage}[b]{0.11\linewidth}
  \centering
  \centerline{\includegraphics[width=1.9cm,height=3.5cm]{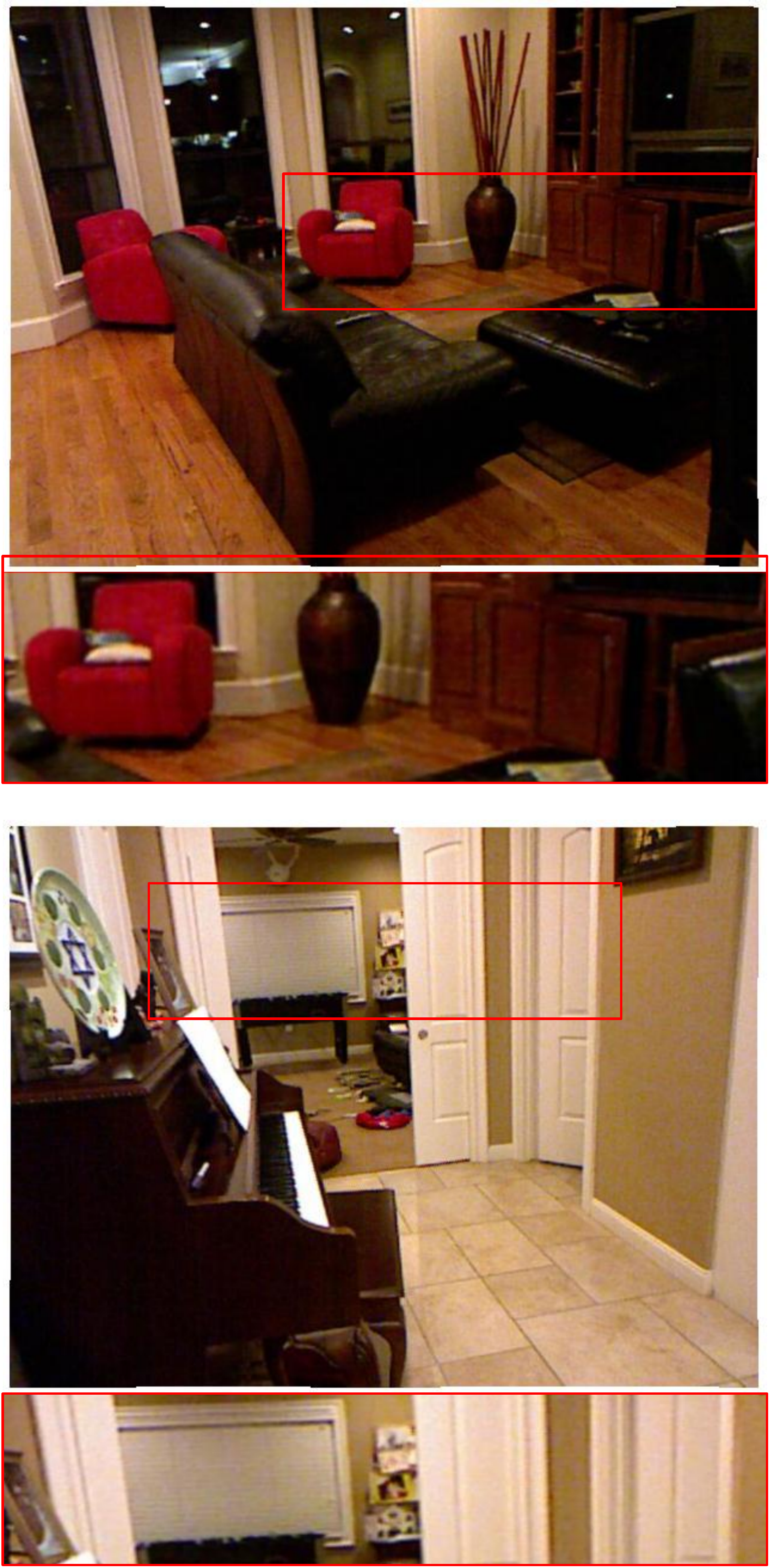}}
  \centerline{(h) GT }\medskip
\end{minipage}
\caption{Results on synthetic hazy images. The details are amplified in the red boxes.}
\label{fig:synthetic}
\end{figure*}

Figure~\ref{fig:synthetic} presents sample results for several synthesized hazy images. It is visible that Meng \etal \cite{Meng2013} tends to over-enhance the inputs and introduces color deviation (\eg, the color of door) while Cai \etal \cite{Cai2016}, Ren \etal \cite{Ren2016} and Li \etal \cite{Li2017allinone} leave haze on the results. In comparison, the results of our DR-Net (ID and IR) are superior to all the others and much more similar to the ground truth data.

In Table~\ref{label:1}, we show the average values of different metrics on testing set. The values in bold represent the best results. Since ground truth images are known for the synthetic images, we use MSE, PSNR (dB), and SSIM metrics for quantitative evaluation on testing set. A higher SSIM value indicates a result that is more close to the ground truth in terms of structural properties. A higher PSNR (lower MSE) value indicates similarity in terms of pixel-wise values.

As visible, our ID and IR achieve the best MSE, PSNR, and SSIM values. The reasons include 
1) ID directly minimizes the reconstruction loss to avoid the accumulated errors from separated parameters estimation; and 2) in the optimization process, SSIM and MSE losses are used.

\begin{table}[htb]
\renewcommand{\arraystretch}{1}
\caption{ Quantitative evaluation on synthetic testing set.}
\centering
\begin{tabular}{clcccccc}
  \hline
 \textbf{Metrics}  & \textbf{\cite{Meng2013}} & \textbf{\cite{Cai2016}} & \textbf{\cite{Ren2016}} & \textbf{\cite{Li2017allinone}} & \textbf{ID} & \textbf{IR}\\
 \hline
MSE         & 2053   & 655   & 1381   & 1048   &  \textbf{477} &\textbf{526} \\
PSNR      &  15.0 & 19.8  &  16.7 & 17.9 &   \textbf{21.3} & \textbf{20.9} \\
SSIM          &  0.78  & 0.87   &  0.80  & 0.81   & \textbf{0.91}& \textbf{0.89}\\
\hline
\label{label:1}
\end{tabular}
\vspace{-3mm}
\end{table}

\subsection{Experiments on Real Data}

\begin{figure*}[htb]
  \centering
\begin{minipage}[b]{0.13\linewidth}
  \centering
  \centerline{\includegraphics[width=2.2cm,height=5cm]{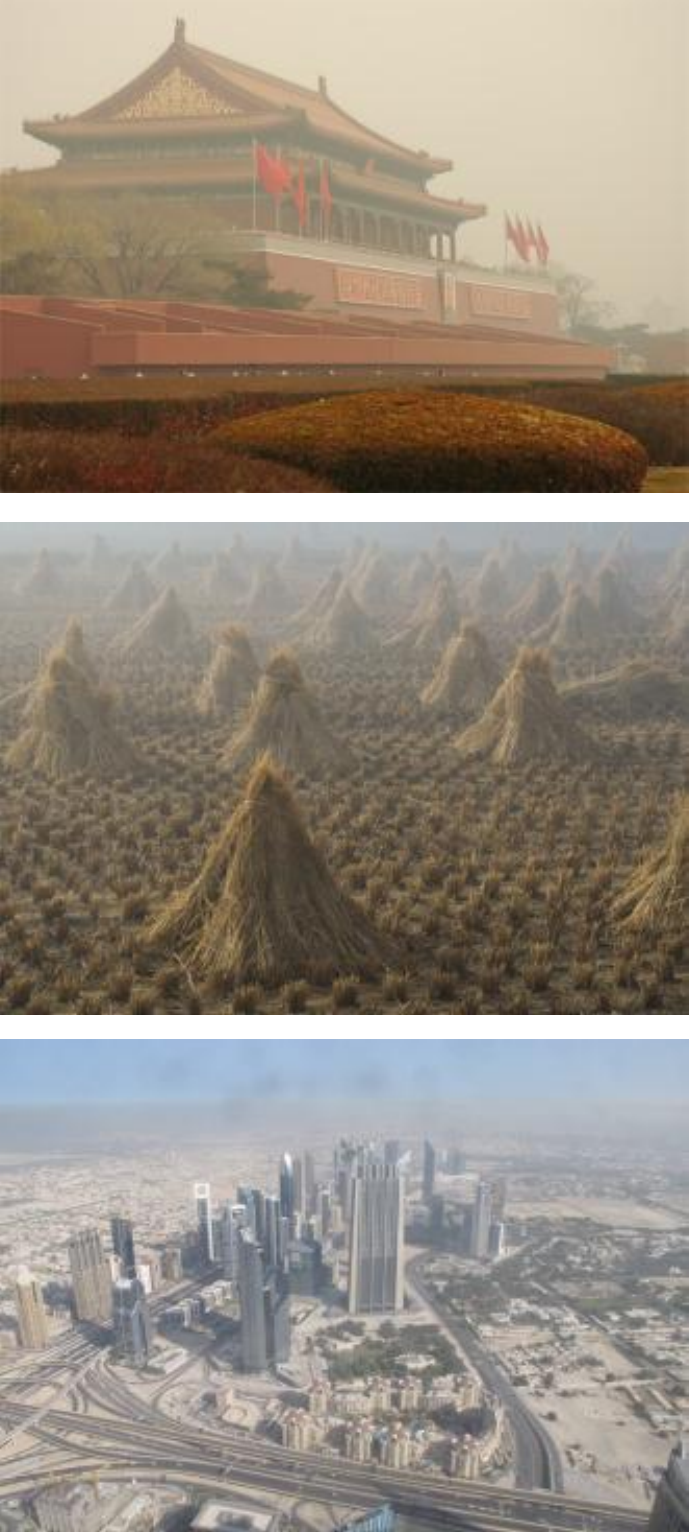}}
  \centerline{(a)Hazy images}\medskip
\end{minipage}
\begin{minipage}[b]{0.13\linewidth}
  \centering
  \centerline{\includegraphics[width=2.2cm,height=5cm]{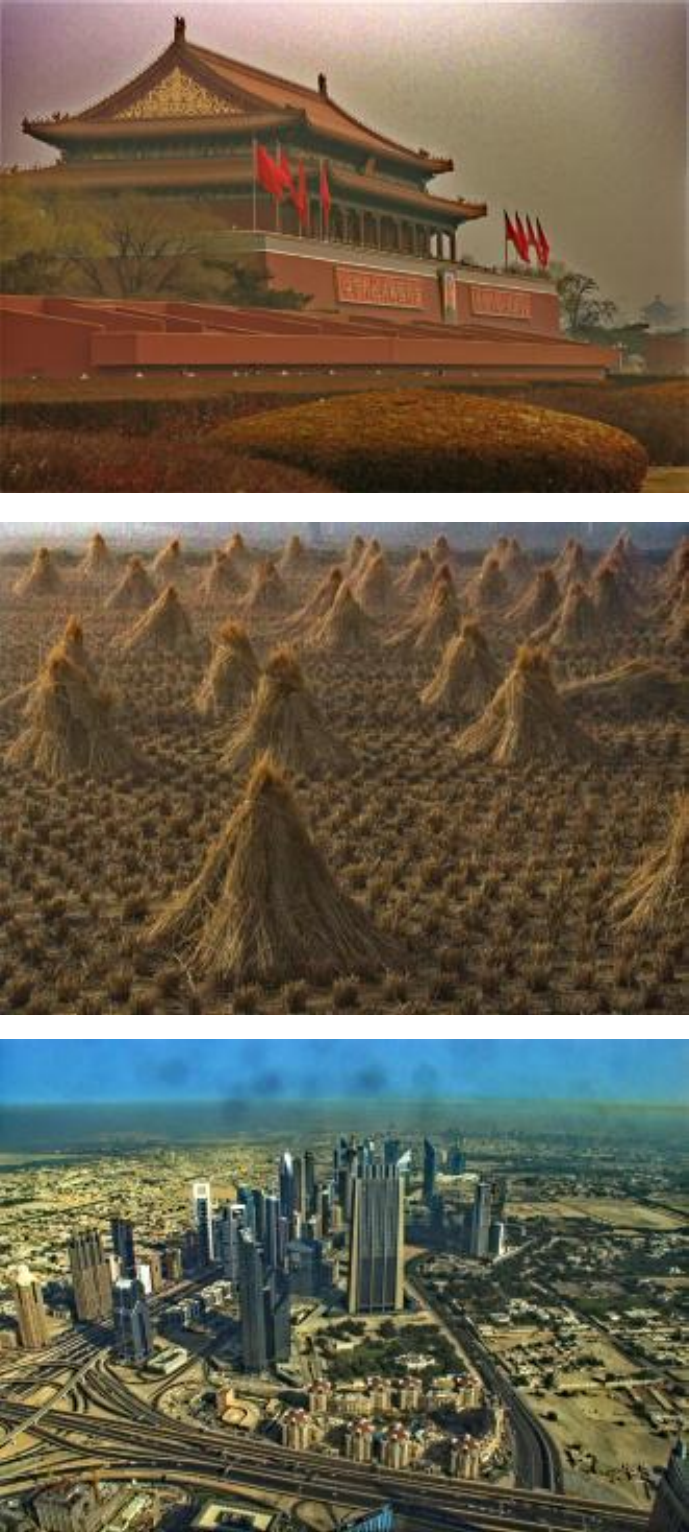}}
  \centerline{(b) \cite{Meng2013}}\medskip
\end{minipage}
\begin{minipage}[b]{0.13\linewidth}
  \centering
  \centerline{\includegraphics[width=2.2cm,height=5cm]{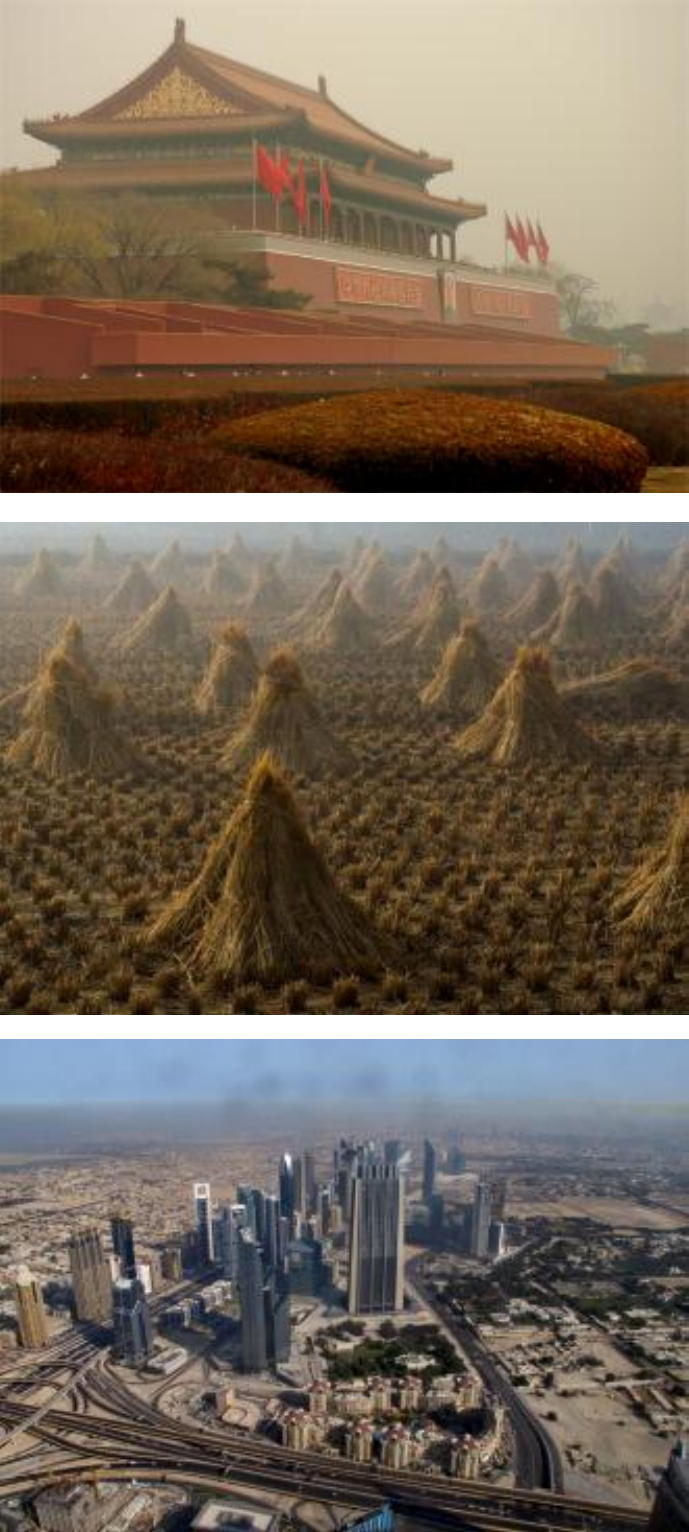}}
  \centerline{(c) \cite{Cai2016}}\medskip
\end{minipage}
\begin{minipage}[b]{0.13\linewidth}
  \centering
  \centerline{\includegraphics[width=2.2cm,height=5cm]{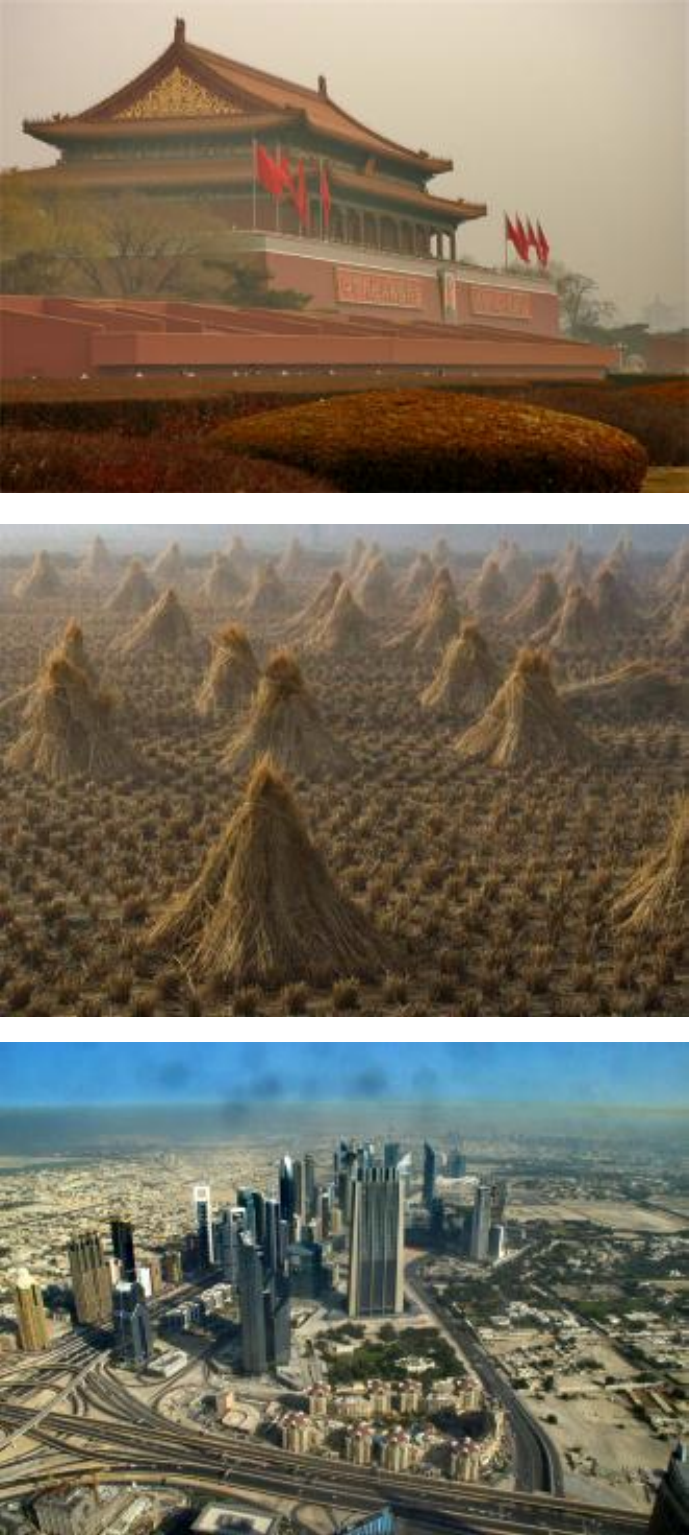}}
  \centerline{(d)  \cite{Ren2016}}\medskip
\end{minipage}
\begin{minipage}[b]{0.13\linewidth}
  \centering
  \centerline{\includegraphics[width=2.2cm,height=5cm]{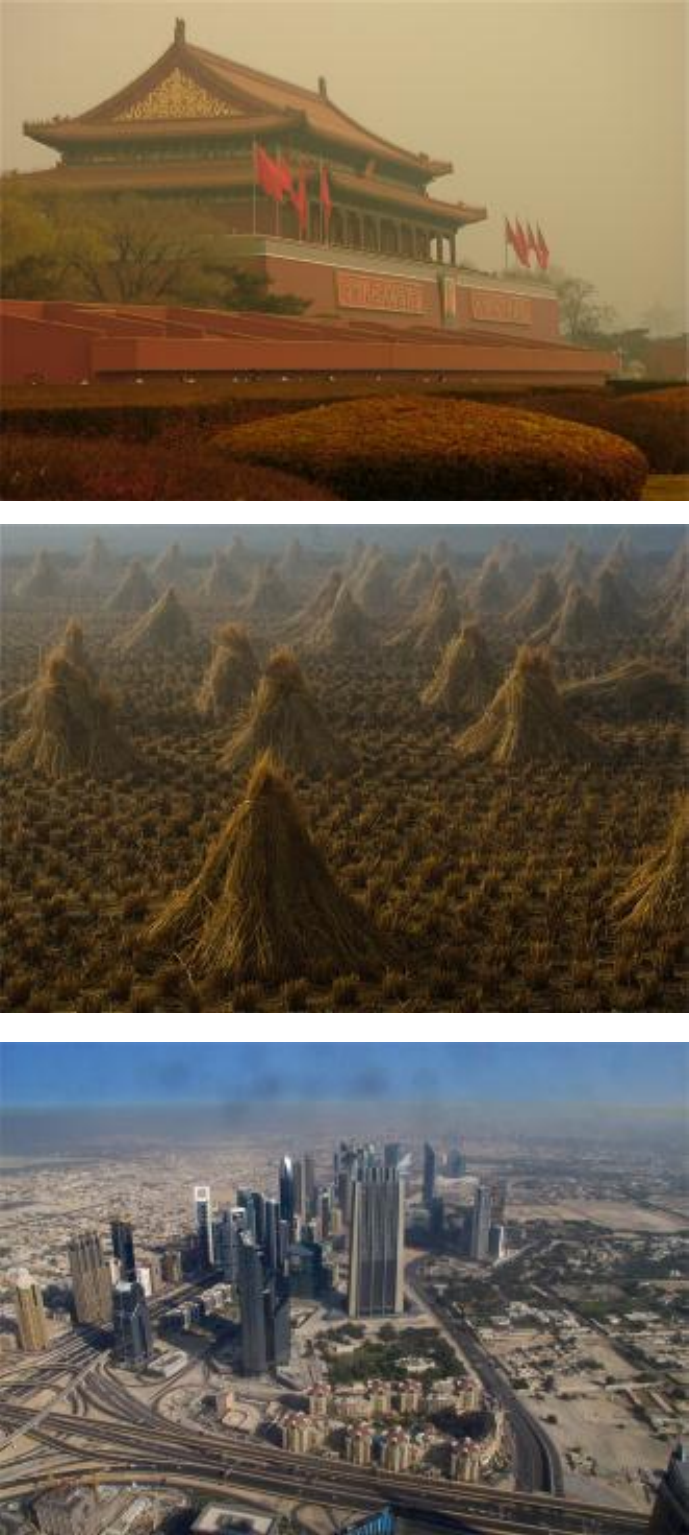}}
  \centerline{(e) \cite{Li2017allinone}}\medskip
\end{minipage}
\begin{minipage}[b]{0.13\linewidth}
  \centering
  \centerline{\includegraphics[width=2.2cm,height=5cm]{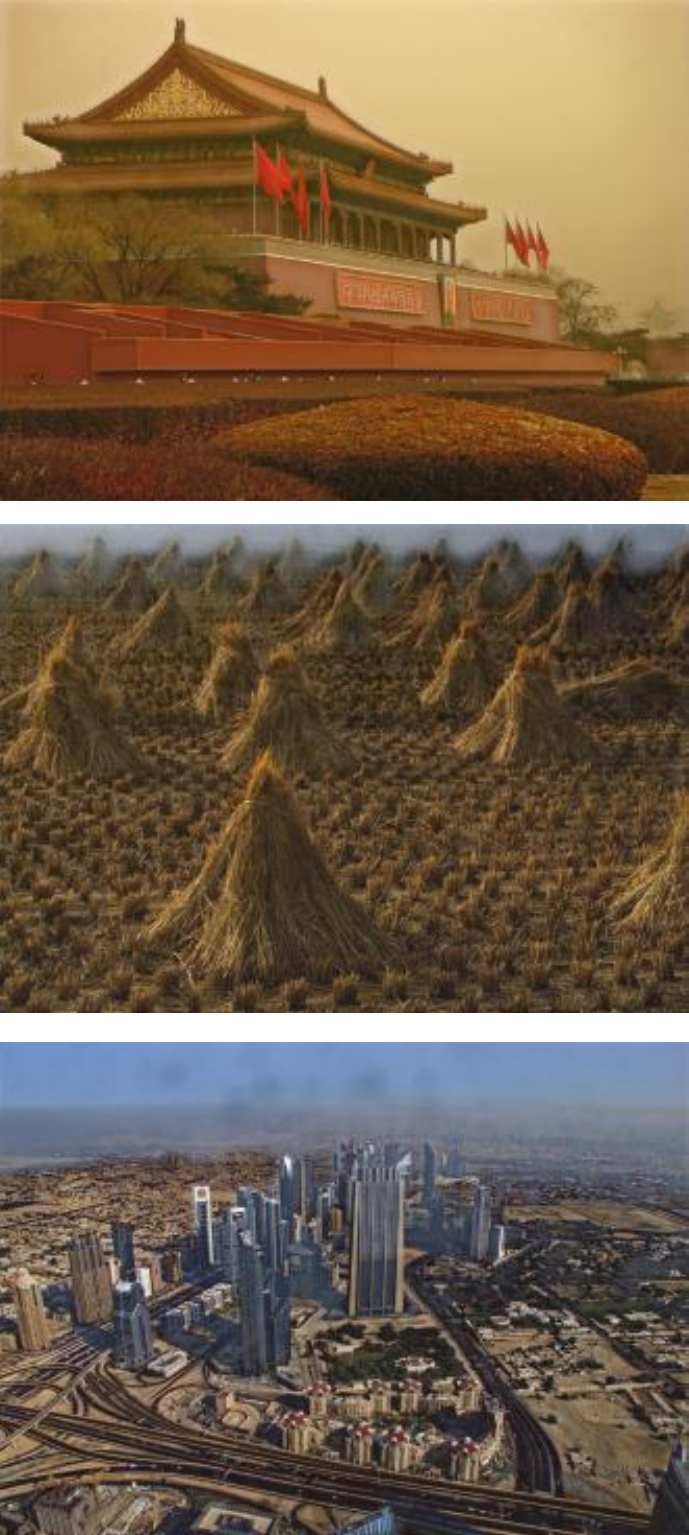}}
  \centerline{(f) ID }\medskip
\end{minipage}
\begin{minipage}[b]{0.13\linewidth}
  \centering
  \centerline{\includegraphics[width=2.2cm,height=5cm]{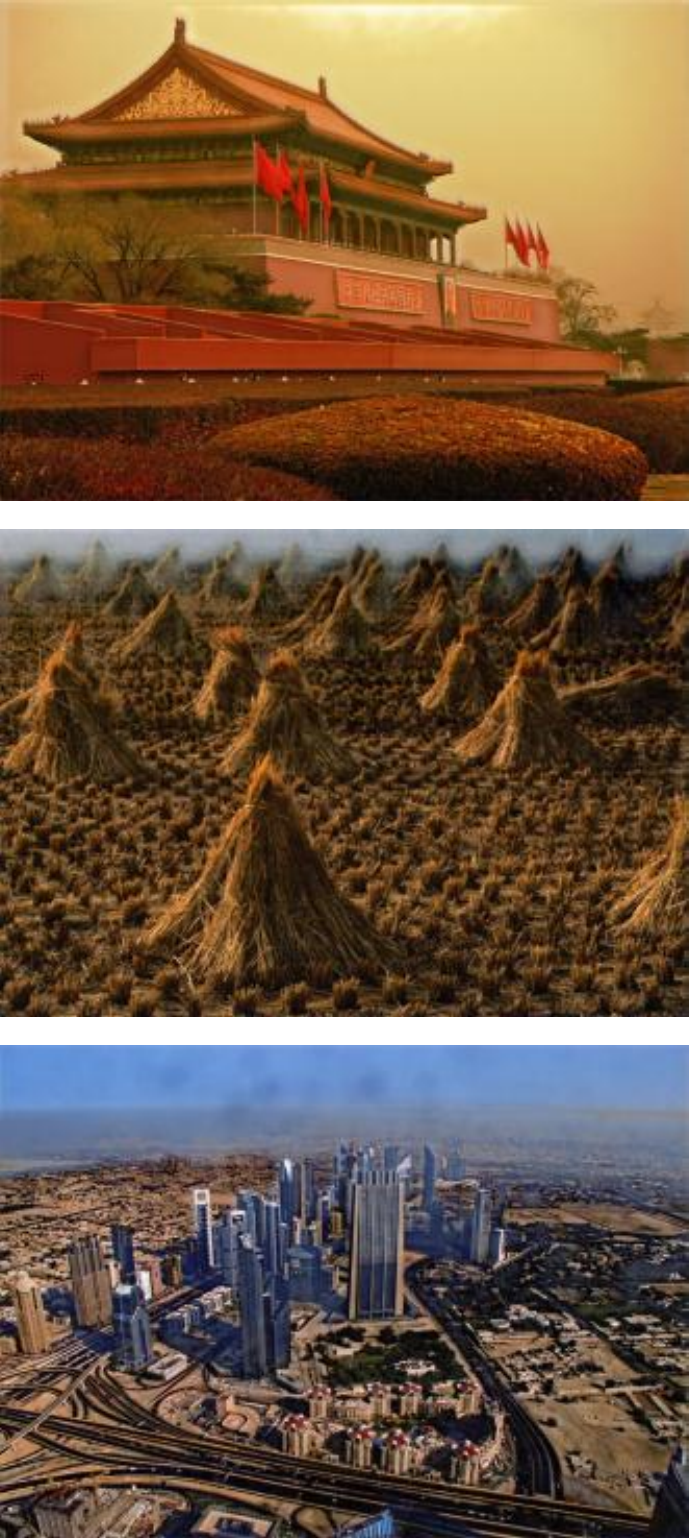}}
  \centerline{(g) IR }\medskip
\end{minipage}
\caption{Results on real hazy images. From top to bottom are `Tiananmen', `wheat', and `Dubai'. }
\label{fig:subjective}
\end{figure*}

To demonstrate the effectiveness of DR-Net on real data, we collected a test dataset including 30 hazy images with a variety of haze levels, image content, and light conditions downloaded from the Internet. Figure~\ref{fig:subjective} shows visual comparisons on several real hazy images.

In Figure~\ref{fig:subjective}, the results of Meng \etal \cite{Meng2013}, ID, and IR leave less haze artifacts. Nevertheless, Meng \etal \cite{Meng2013} introduces obvious color deviation (\eg, the background of `Tiananmen' and `Dubai'). The results of Cai \etal \cite{Cai2016}, Ren \etal \cite{Ren2016} and Li \etal \cite{Li2017allinone} remain haze (\eg, the background of `wheat' ). Compared to other methods, our IR has the most competitive visual results among all, with high contrast, clear structure, plausible details and vivid color.

Since the ground truth images are not available for the real data, we conduct a user study to provide realistic feedback for quantitative evaluation. We invited 10 subjects who had experience with image processing to rank the visual quality. Subjects were allowed to zoom in and out at will without time restriction. The scores range from 1 (worst) to 10 (best). We collected $10\times 30=300$ scores and presented the average scores of different methods in Table~\ref{label:2}. 
\begin{table}[htb]
\renewcommand{\arraystretch}{1}
\caption{ User study on 30 real hazy images.}
\centering
\begin{tabular}{clcccccc}
  \hline
 \textbf{Metric}  & \textbf{\cite{Meng2013}} & \textbf{\cite{Cai2016}} & \textbf{\cite{Ren2016}} & \textbf{\cite{Li2017allinone}} & \textbf{ID}& \textbf{IR}\\
 \hline
Scores  & 4.7 & 6.1& 7.7 & 6.5 & \textbf{8.2} & \textbf{9.1} \\
\hline
\label{label:2}
\end{tabular}
\vspace{-3mm}
\end{table}

As shown, our ID and IR receive the best scores, which indicates that, from a visual perspective, our method can produce much better performance on real hazy images. Meng \etal \cite{Meng2013} achieves worst score, which indicates that over-enhancement and color deviation importantly affect the visual quality.
Besides, we believe that there is a gap between real hazy images and synthetic haze images, which leads to IR ranking second in the quantity comparisons on synthetic testing set.

\subsection{Experiments on Challenging Data}

\begin{figure*}[htb]
  \centering
\begin{minipage}[b]{0.13\linewidth}
  \centering
  \centerline{\includegraphics[width=2.2cm,height=5.5cm]{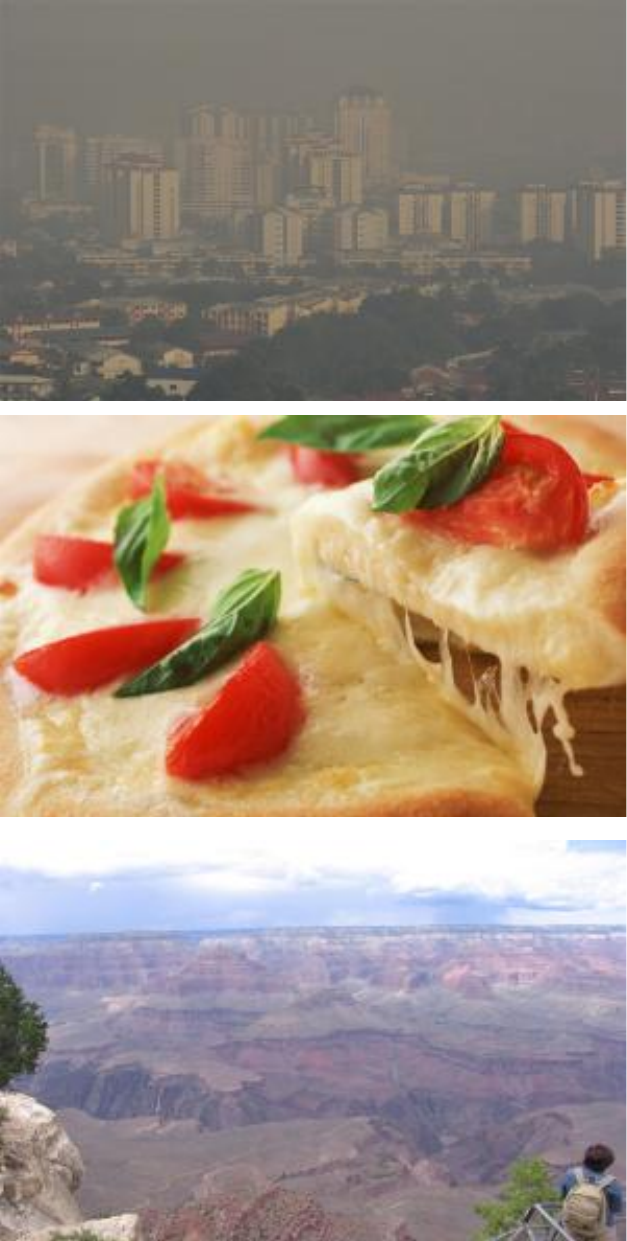}}
  \centerline{(a)Test images}\medskip
\end{minipage}
\begin{minipage}[b]{0.13\linewidth}
  \centering
  \centerline{\includegraphics[width=2.2cm,height=5.5cm]{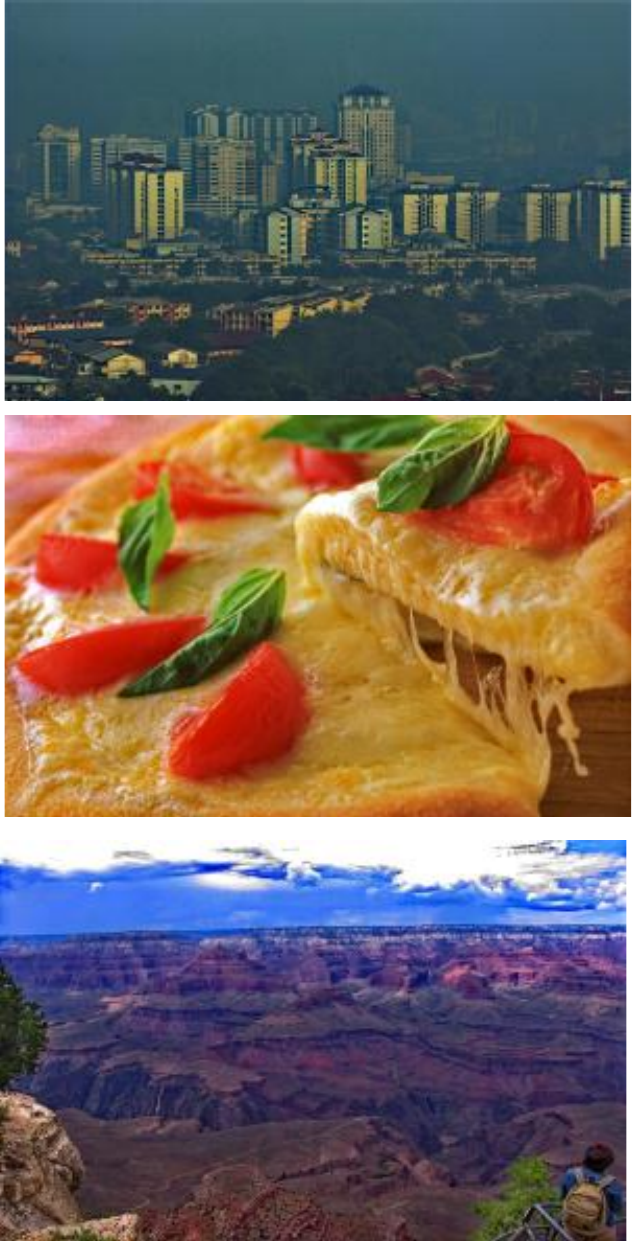}}
  \centerline{(b) \cite{Meng2013}}\medskip
\end{minipage}
\begin{minipage}[b]{0.13\linewidth}
  \centering
  \centerline{\includegraphics[width=2.2cm,height=5.5cm]{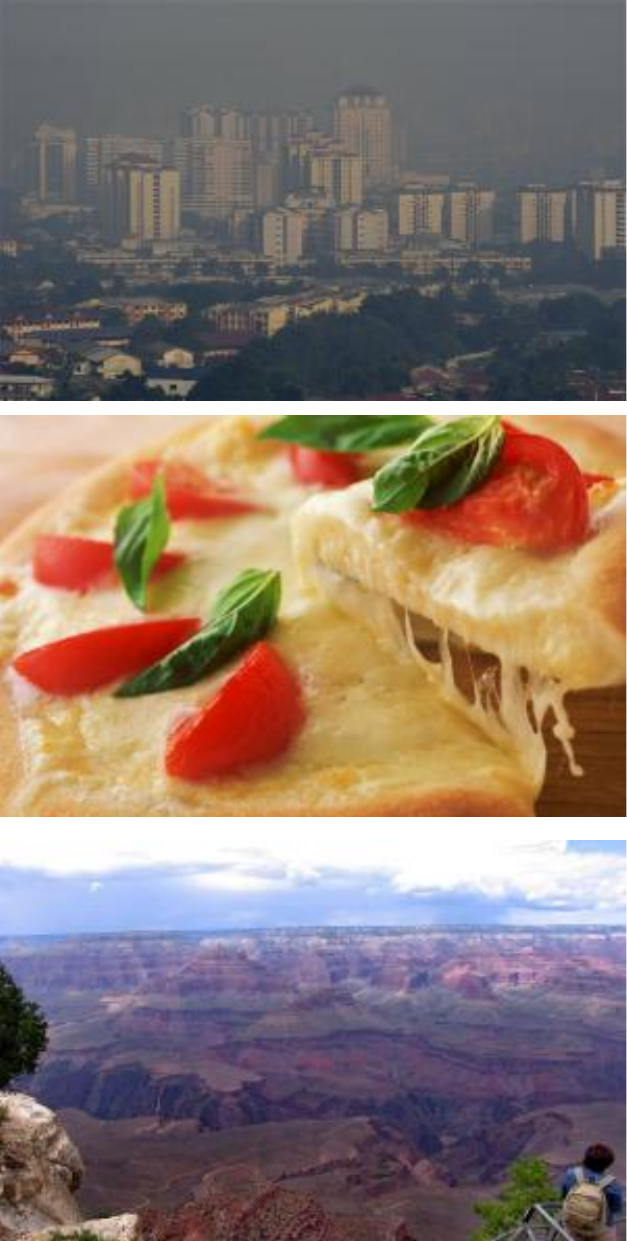}}
  \centerline{(c) \cite{Cai2016}}\medskip
\end{minipage}
\begin{minipage}[b]{0.13\linewidth}
  \centering
  \centerline{\includegraphics[width=2.2cm,height=5.5cm]{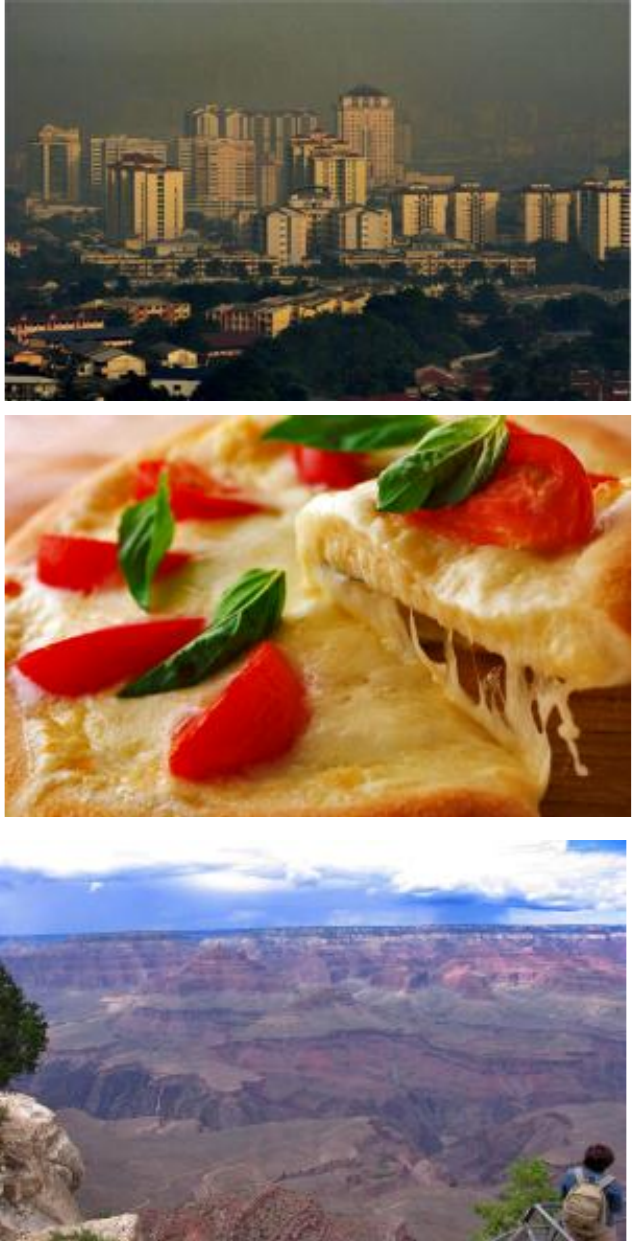}}
  \centerline{(d)  \cite{Ren2016}}\medskip
\end{minipage}
\begin{minipage}[b]{0.13\linewidth}
  \centering
  \centerline{\includegraphics[width=2.2cm,height=5.5cm]{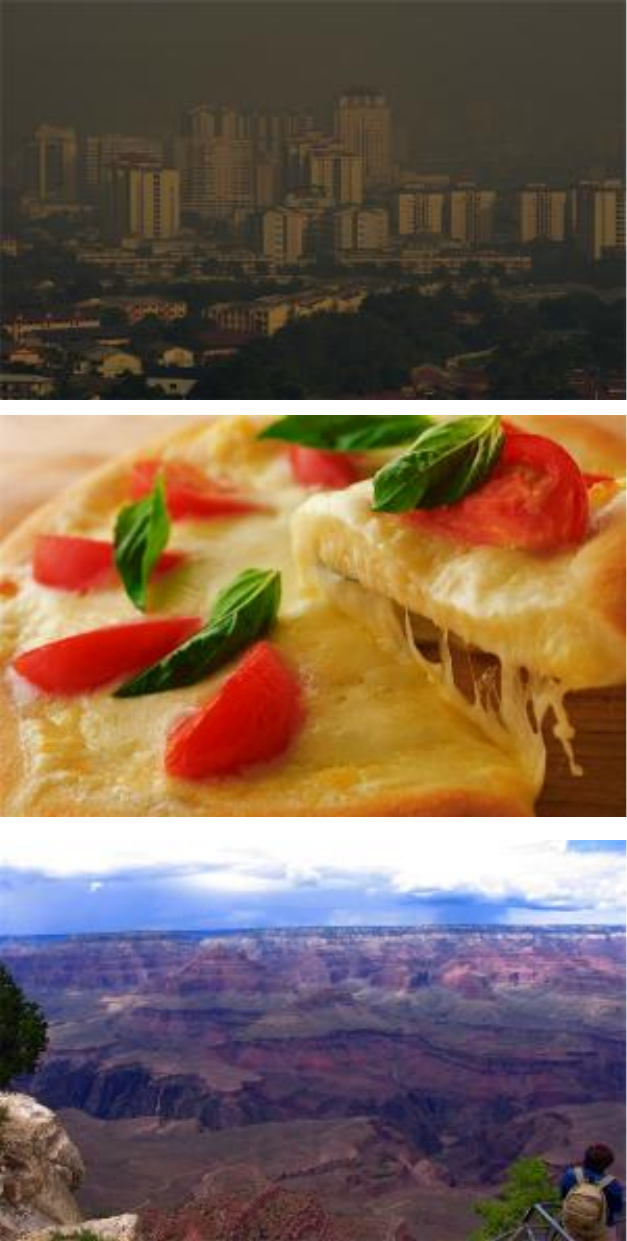}}
  \centerline{(e) \cite{Li2017allinone}}\medskip
\end{minipage}
\begin{minipage}[b]{0.13\linewidth}
  \centering
  \centerline{\includegraphics[width=2.2cm,height=5.5cm]{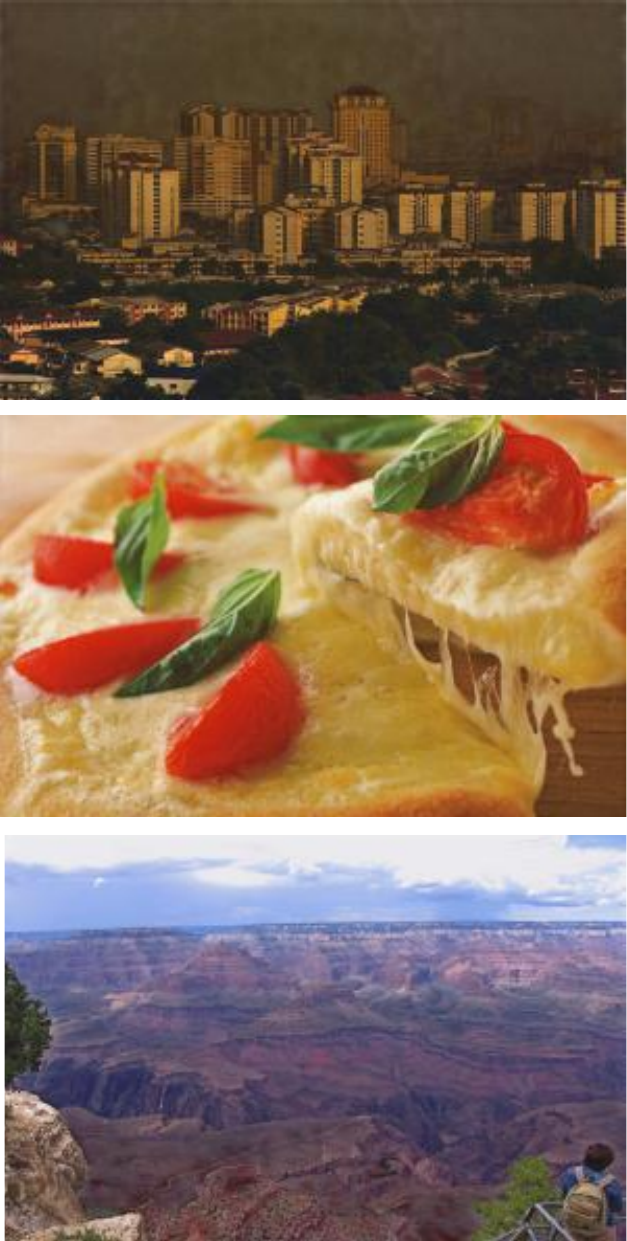}}
  \centerline{(f) ID}\medskip
\end{minipage}
\begin{minipage}[b]{0.13\linewidth}
  \centering
  \centerline{\includegraphics[width=2.2cm,height=5.5cm]{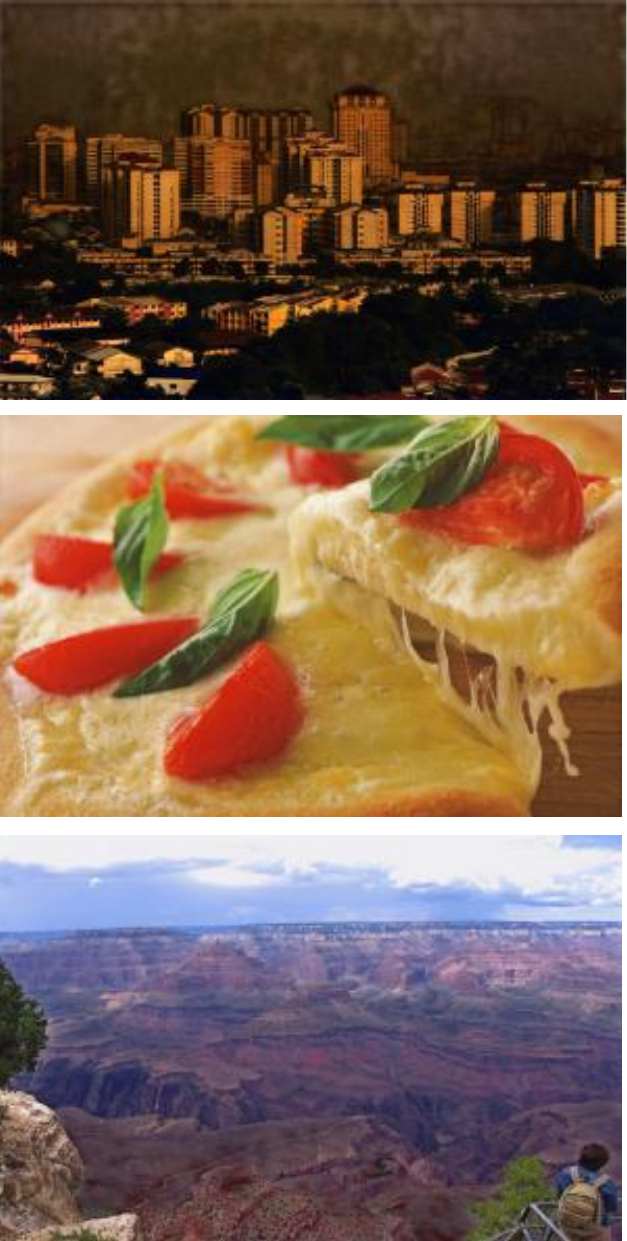}}
  \centerline{(g) IR }\medskip
\end{minipage}
\caption{Results on challenging images. From top to bottom are `buildings' with low light and heavy haze, `pizza' without haze and `canyon' with white cloud. }
\label{fig:challenging}
\end{figure*}

We expect that dehazing methods also generalize well to challenging data. Thus, we carry out an experiment on challenging data, including image with heavy haze, haze-free image, and image with white object. The visual comparisons are shown in Figure~\ref{fig:challenging}.

In Figure~\ref{fig:challenging}, for image `buildings', the results of Ren \etal \cite{Ren2016}, ID and IR remain less haze. Besides, the buildings in the IR has better details and brightness. Other methods, however, have less effect on image `buildings', even introducing bluish color deviation. Generally, it is difficult to recover images with heavy haze. For image `pizza', Meng \etal \cite{Meng2013}, Ren \etal \cite{Ren2016} and Li \etal \cite{Li2017allinone} produce over-enhancement (\eg, the color of pizza). In practice, it is expected that image dehazing methods have less effect on haze-free image. For image `canyon', all of methods are unsensitive to white cloud. However, the results of Meng \etal \cite{Meng2013} significantly suffers from color deviation on the canyon. Although DR-Net is trained using synthesized data, it generalizes well to outdoor hazy images even captured under challenging scenes. Furthermore, different from Ren \etal \cite{Ren2016} which requires different gamma values to amend transmission value according to haze concentrations, our DR-Net does not need manual parameters tuning, which is desired for practical applications.

\subsection{Applications}
To illustrate the potential usage of our DR-Net, we employ it on several computer vision tasks such as keypoint matching and object detection and recognition.

\textbf{Keypoint matching} is the base of many important computer vision problems. We employ the SURF operator \cite{Bay2006} for original haze image pair and our refined image pair.

\begin{figure}[htb]
  \centering
\begin{minipage}[b]{0.3\linewidth}
  \centering
  \centerline{\includegraphics[width=8cm,height=2.3cm]{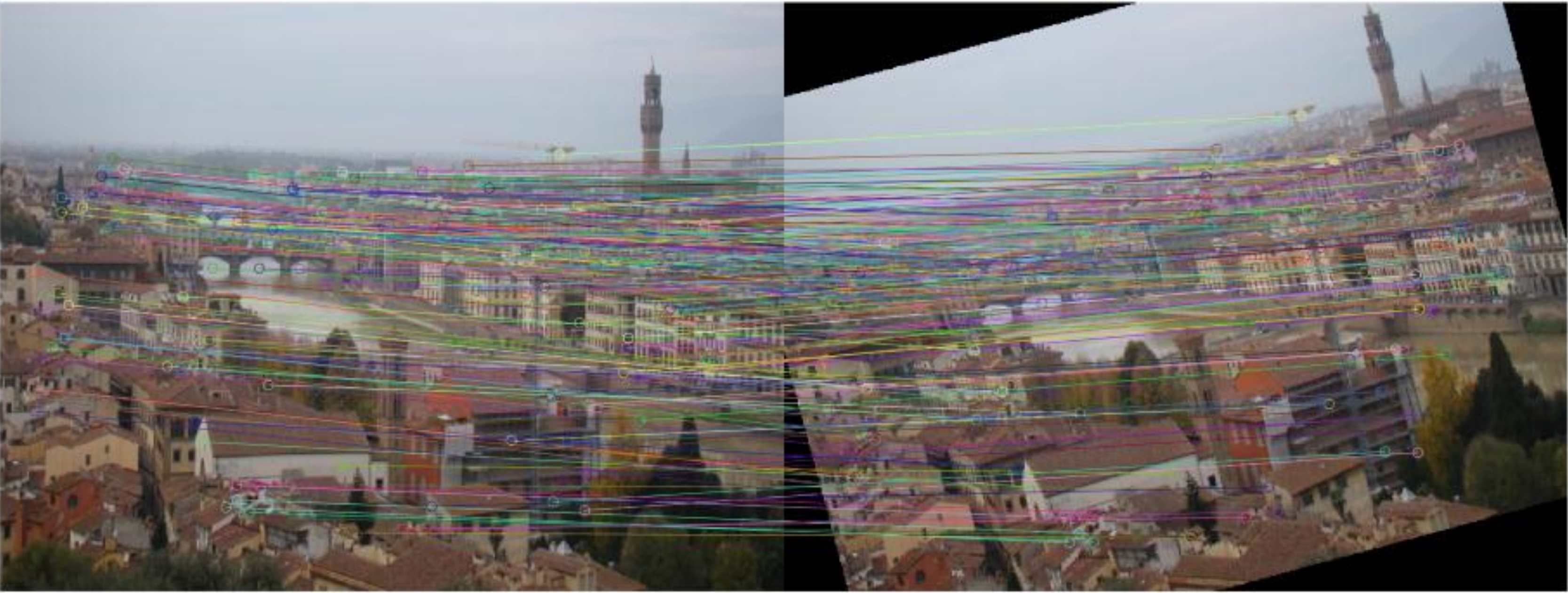}}
  \centerline{(a)Hazy image pair}\medskip
\end{minipage}

\begin{minipage}[b]{0.3\linewidth}
  \centering
  \centerline{\includegraphics[width=8cm,height=2.3cm]{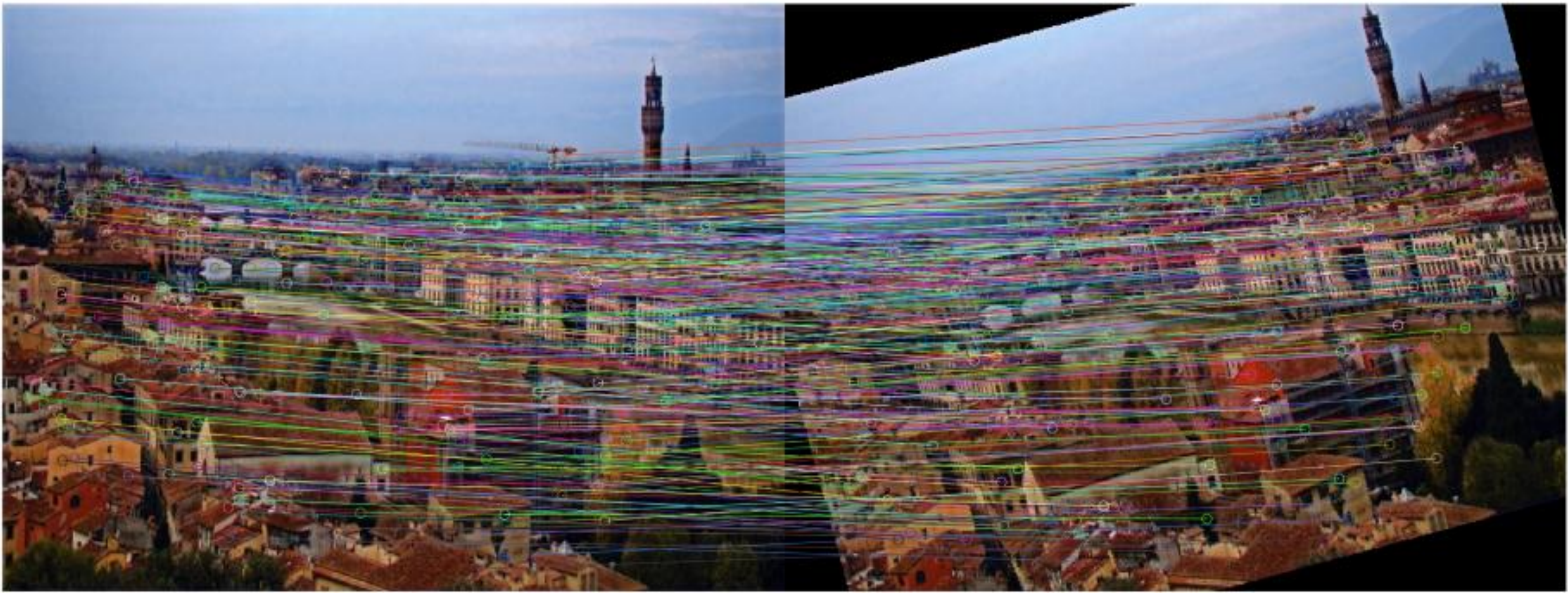}}
  \centerline{(b)Refined result pair}\medskip
\end{minipage}
\caption{An example of keypoint matching. }
\label{fig:surf}
\end{figure}

In Figure~\ref{fig:surf}, the initial matches are 226 while the matches of our refined pair are 344. DR-Net significantly increases the number of SURF keypoint matching.

\begin{figure}[htb]
  \centering
\begin{minipage}[b]{0.45\linewidth}
  \centering
  \centerline{\includegraphics[width=3cm,height=1.9cm]{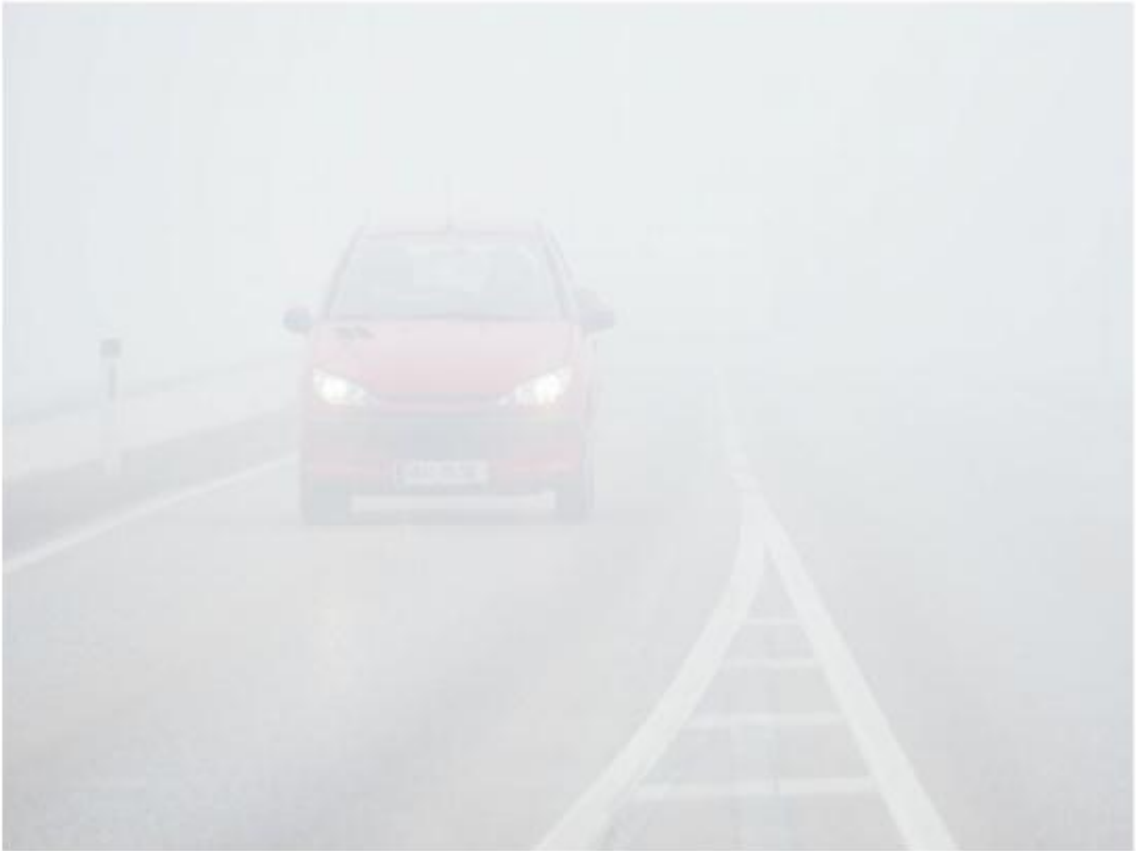}}
  \centerline{(a)Hazy image}\medskip
\end{minipage}
\begin{minipage}[b]{0.45\linewidth}
  \centering
  \centerline{\includegraphics[width=3cm,height=1.9cm]{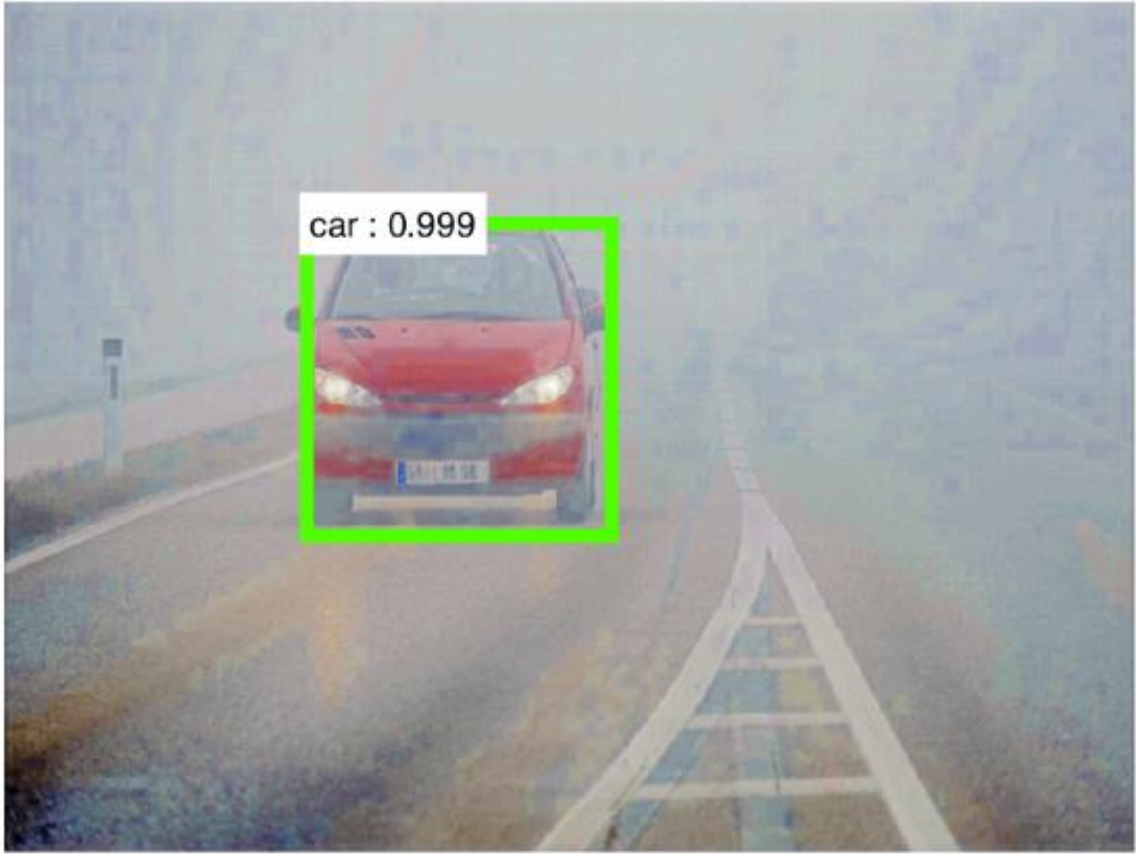}}
  \centerline{(b)Refined result }\medskip
\end{minipage}
\caption{An example of object detection and recognition. }
\label{fig:frcnn}
\end{figure}

\textbf{Object detection and recognition} has attracted much attention recently. We use the fast R-CNN \cite{Girshick2015} trained on VOC 2007 dataset to valid the performance of DR-Net.
In Figure~\ref{fig:frcnn}, the image after processing by DR-Net improves the accuracy of objects detection and recognition. Additionally, due to the low quality of input image with JPG compression, our method produces artifacts. This phenomenon also can be observed in the results of other image dehazing methods. In future work, we will investigate how to reduce the effects induced by low quality images.

In short, experiments on applications provide additional evidence for the potential usage of DR-Net.

\subsection{Effects of Transmission Prediction Subnetwork}

To analyze the effects of the transmission prediction subnetwork, we fix the default parameter settings, and then remove the transmission prediction subnetwork. It took about 15 hours to optimize this DR-Net. Obviously, it took more time than our original DR-Net optimization, which demonstrates that the predicted transmission map is helpful for the training convergence. Besides, we also observed the predicted transmission map has positive effects on the results.

\begin{figure}[htb]
  \centering
\begin{minipage}[b]{0.32\linewidth}
  \centering
  \centerline{\includegraphics[width=2.0cm,height=1.5cm]{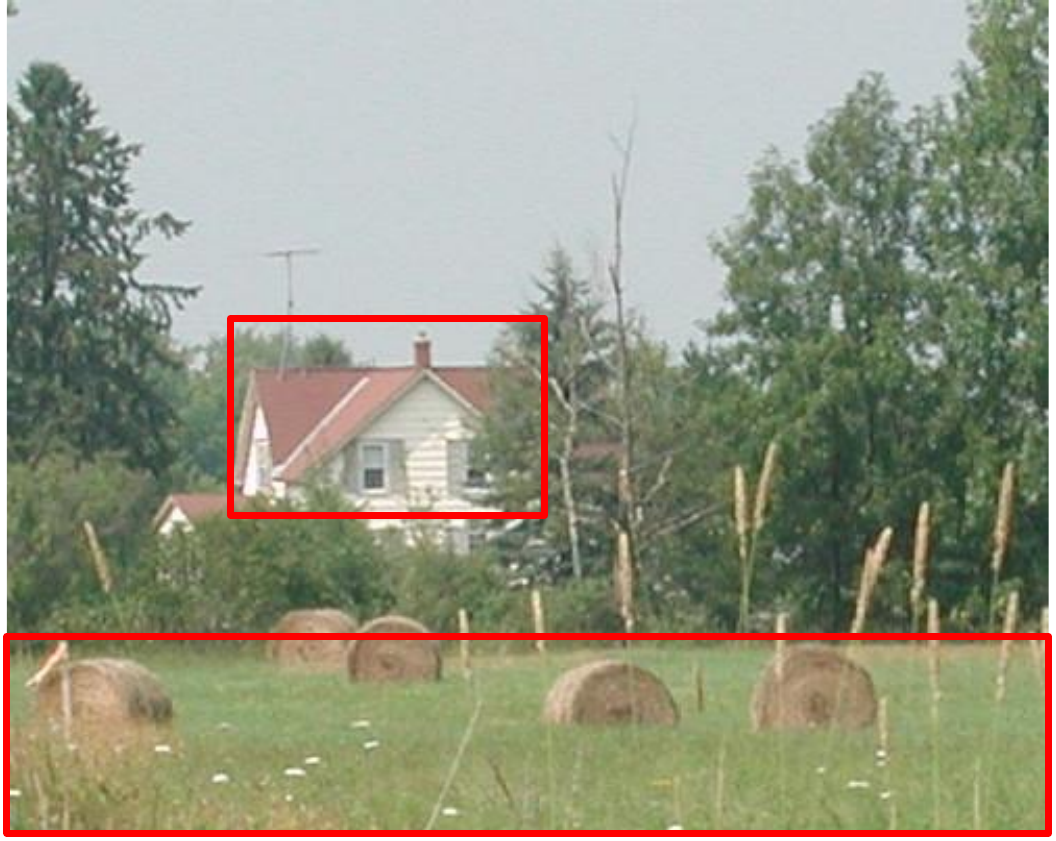}}
  \centerline{(a)}\medskip
\end{minipage}
\begin{minipage}[b]{0.32\linewidth}
  \centering
  \centerline{\includegraphics[width=2.0cm,height=1.5cm]{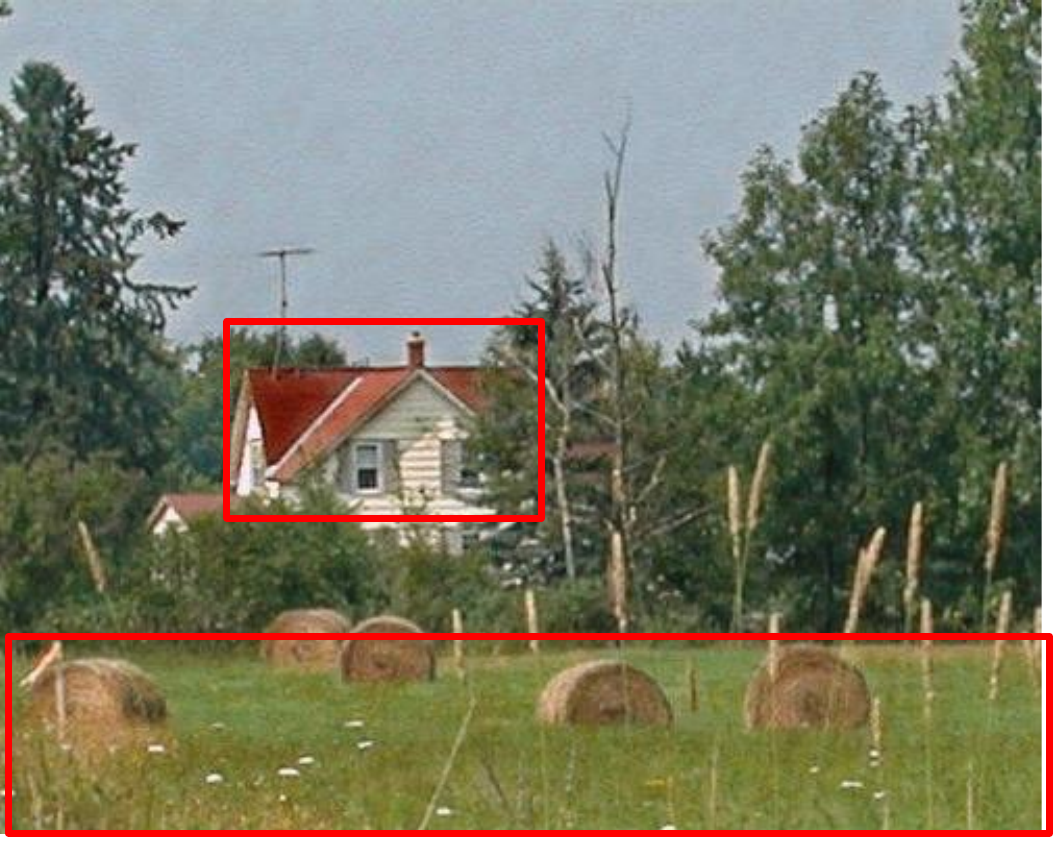}}
  \centerline{(b)}\medskip
\end{minipage}
\begin{minipage}[b]{0.32\linewidth}
  \centering
  \centerline{\includegraphics[width=2.0cm,height=1.5cm]{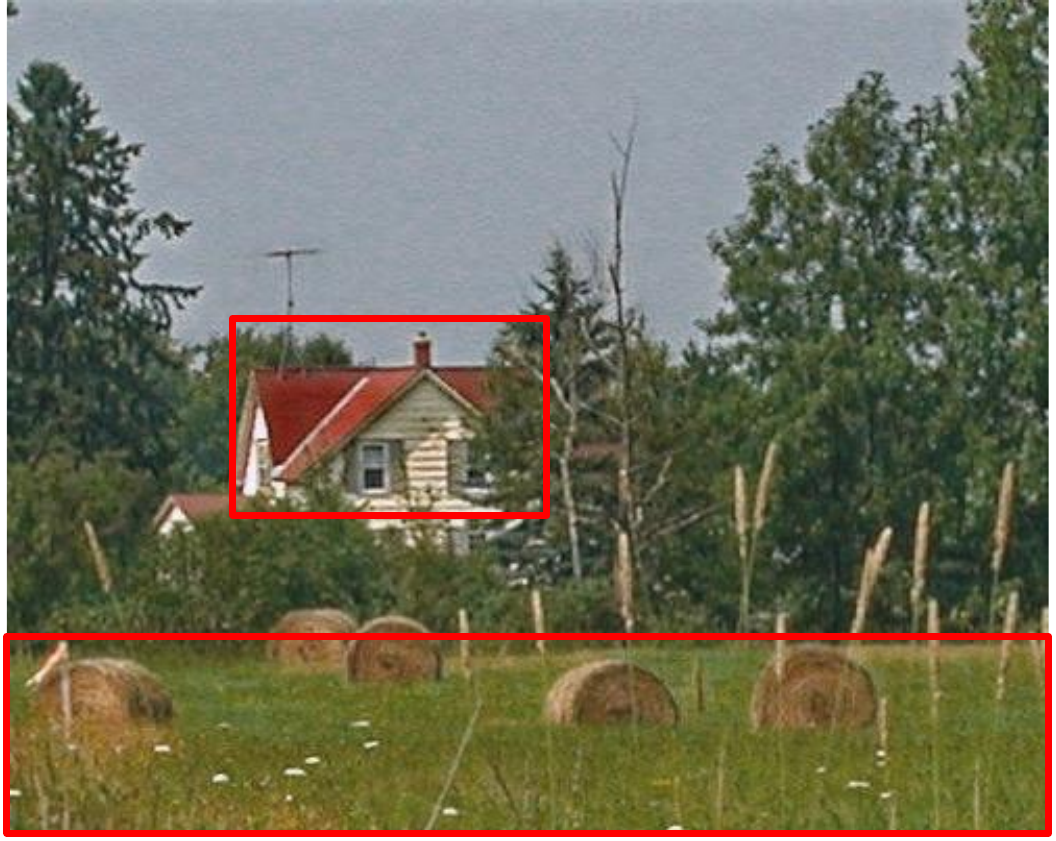}}
  \centerline{(c)}\medskip
\end{minipage}

\begin{minipage}[b]{0.32\linewidth}
  \centering
  \centerline{\includegraphics[width=2.0cm,height=1.5cm]{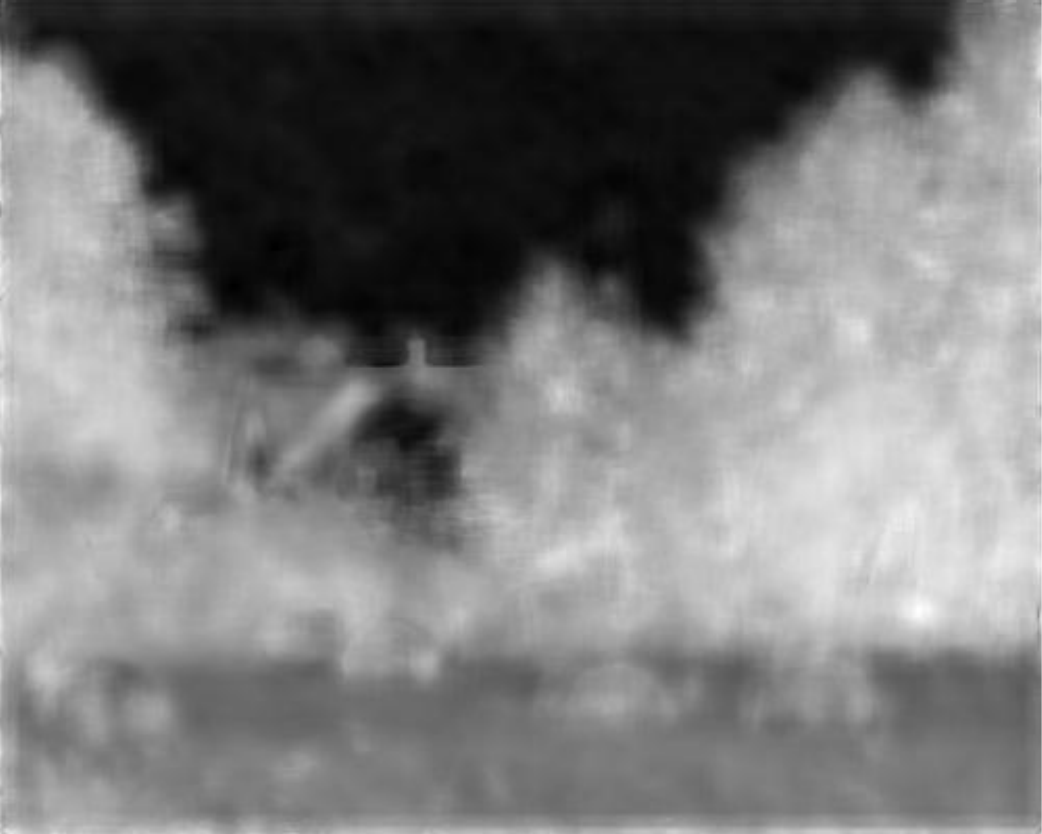}}
  \centerline{(d) }\medskip
\end{minipage}
\begin{minipage}[b]{0.32\linewidth}
  \centering
  \centerline{\includegraphics[width=2.0cm,height=1.5cm]{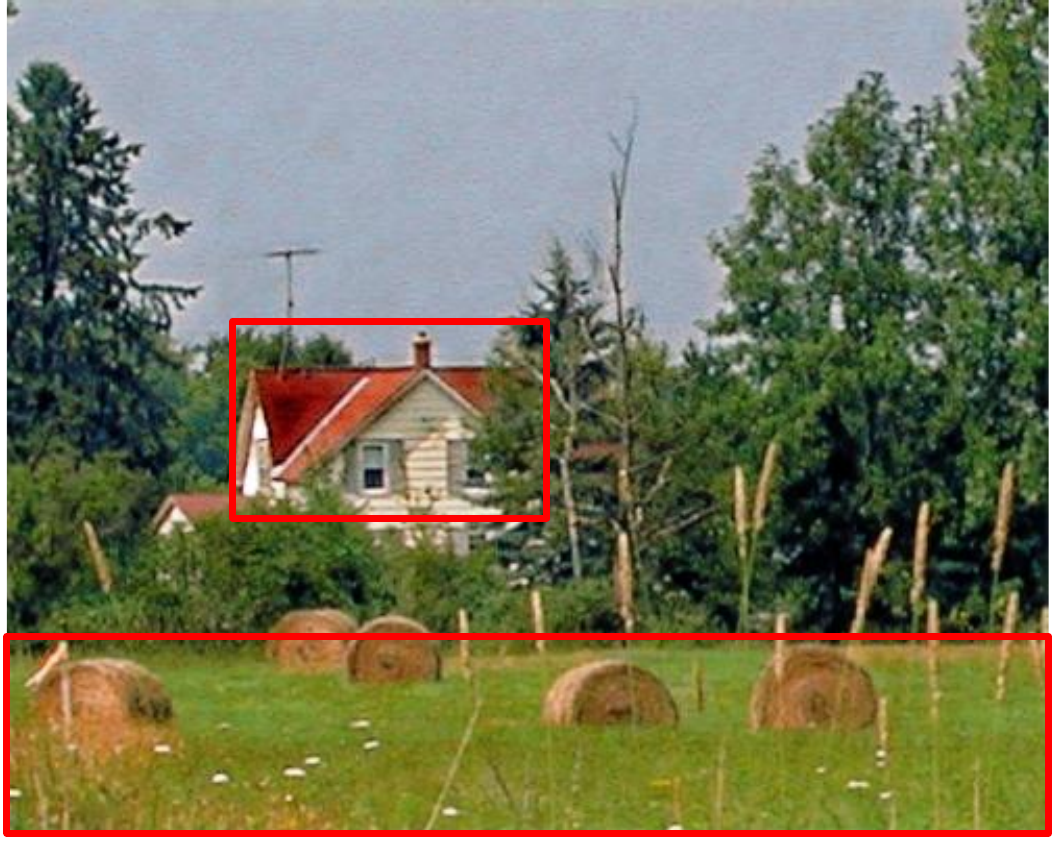}}
  \centerline{(e) }\medskip
\end{minipage}
\begin{minipage}[b]{0.32\linewidth}
  \centering
  \centerline{\includegraphics[width=2.0cm,height=1.5cm]{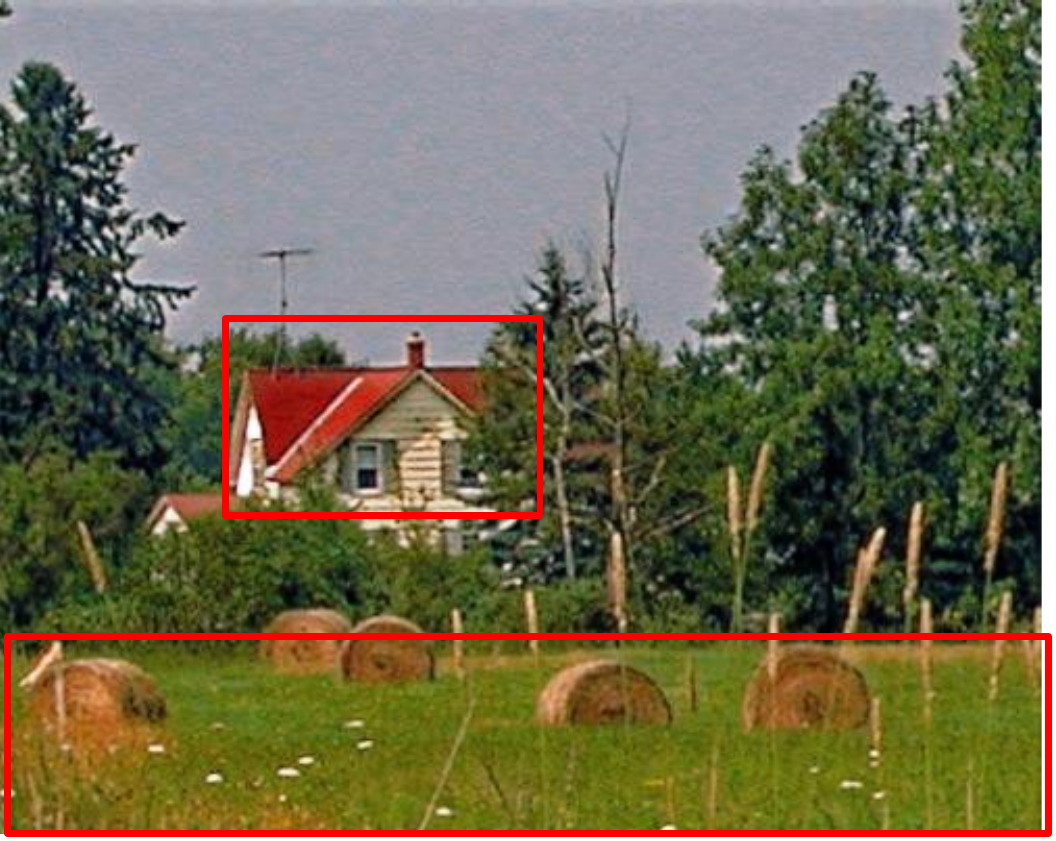}}
 \centerline{(f) }\medskip
\end{minipage}
\caption{An example to illustrate the effects of transmission prediction subnetwork. (a) Hazy image. (b) Dehazed result from dehazing subnetworks without transmission prediction subnetwork. (c) Refined result of (b).  (d) The predicted transmission by transmission prediction subnetwork. (e) Dehazed result from dehazing subnetworks with transmission prediction subnetwork. (f) Refined result of (e). (Best viewed on high-resolution display with zoom-in.)}
\label{fig:transmission}
\end{figure}

In Figure~\ref{fig:transmission}, the hazed and refined results from DR-Net with transmission prediction subnetwork has superior visual quality than the one without transmission prediction subnetwork (\eg, the regions of grassland and roof), which indicates the positive effects of the predicted transmission map.
\section{Conclusion}
We have presented a deep learning model for single image dehazing, namely DR-Net, which includes three subnetworks. For haze removal, we propose a transmission prediction subnetwork and a haze removal subnetwork, which are designed to reconstruct clear image steered by transmission map.  To improve the contrast and color of the dehazed result, we propose a refinement subnetwork based on weakly supervised learning of image-to-image translation. Experimental results on synthetic and real data have demonstrated that DR-Net achieves state-of-the-art performance.

\ifCLASSOPTIONcaptionsoff
  \newpage
\fi

\end{document}